\documentclass[10pt,twocolumn,letterpaper]{article}

\usepackage[pagenumbers]{cvpr} 

\definecolor{cvprblue}{rgb}{0.21,0.49,0.74}
\usepackage[pagebackref,breaklinks,colorlinks,allcolors=cvprblue]{hyperref}

\usepackage{booktabs}
\usepackage{multirow}
\usepackage{multicol}
\usepackage{xcolor}
\definecolor{darkgreen}{RGB}{0,128,0} 
\definecolor{lightblue}{rgb}{0.9,0.95,1.0} 
\definecolor{lightgray}{RGB}{240, 243, 246}
\usepackage{colortbl}
\usepackage{pifont}

\definecolor{navy}{RGB}{0, 0, 128}
\definecolor{brickred}{RGB}{203, 65, 84}
\definecolor{darkgreen}{RGB}{0, 100, 0}

\newcommand{\cmark}{\textcolor{darkgreen}{\ding{51}}}  
\newcommand{\xmark}{\textcolor{red}{\ding{55}}}

\usepackage{graphicx}
\usepackage{array}
\usepackage[most]{tcolorbox}
\usepackage{fancyvrb}

\newcommand{\ours}{\textbf{MemRU}\xspace}

\def\confName{CVPR}
\def\confYear{2027}

\title{Read Before You Use: \\ Improving Visual Memory Utilization in Vision-Language Models}

\renewcommand{\thefootnote}{\fnsymbol{footnote}}

\author{Junseok Kim$^{1}$,
Kyungmin Min$^{1}$,
Kyomin Jung$^{1,2}$\footnotemark[2] \\
$^{1}$IPAI, Seoul National University \\
$^{2}$Dept. of ECE, Seoul National University \\
Project Page: \href{https://memru-proj.github.io}{https://memru-proj.github.io}
}

\begin{document}


\twocolumn[{%
    \renewcommand\twocolumn[1][]{#1}%
    \vspace{-2em}
    \maketitle
    \begin{center}
        \centering
        \vspace{-0.2in}
        \includegraphics[width=\textwidth]{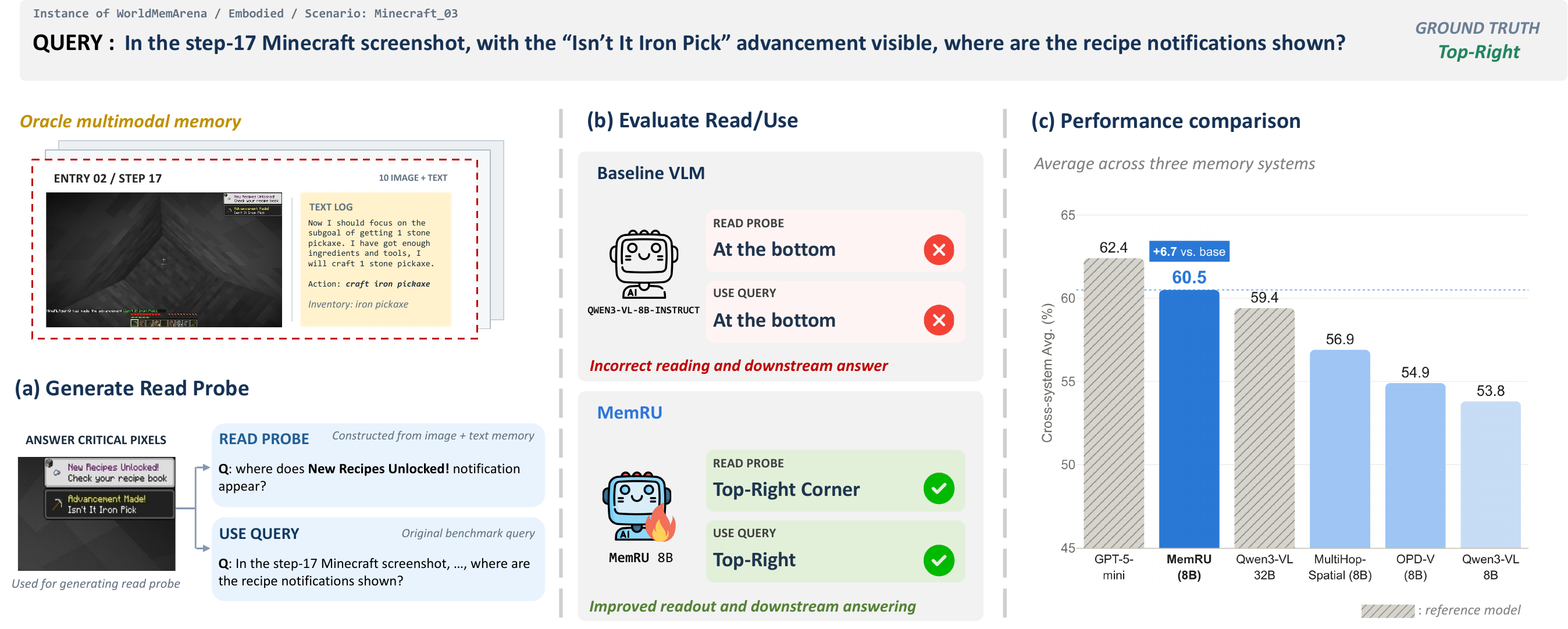}
        \captionof{figure}{
        \textbf{From multimodal memory reading failures to effective utilization.}
        (a) Given multimodal memory with sufficient visual and textual evidence, an image read probe is constructed to assess whether VLM can accurately read relevant visual information.
        (b) While the baseline VLM fails both the read probe and the original QA task, our proposed framework (\ours) correctly answers both.
        (c) Averaged across three memory systems (AUGUSTUS~\cite{jain2025augustus}, Omni-SimpleMem~\cite{liu2026omni}, and UniversalRAG~\cite{yeo2026universalrag}), \ours achieves a 6.7 percentage-point improvement over its 8B base model, outperforming Qwen3-VL-32B and achieving competitive performance with GPT-5-mini.
        }
        \label{fig:teaser}
        \vspace{-1pt}
    \end{center}
}]
\footnotetext[2]{Corresponding author.} 
\renewcommand{\thefootnote}{\arabic{footnote}}
\begin{abstract}
Multimodal memory enables long-horizon vision-language agents to preserve and retrieve past experiences.
While prior research has focused on memory construction and retrieval, effectively using the provided memory remains a challenge for vision-language models (VLMs).
We find that VLMs often fail to answer downstream questions even when sufficient multimodal evidence is provided.
To investigate why VLMs struggle, we replace memory images with reference captions and observe substantial performance gains, suggesting difficulties in reading visual evidence directly.
However, model-generated captions may omit critical information, limiting this approach in practice.
These limitations motivate a closer examination of memory utilization, revealing failures in both evidence reading and downstream question answering.
Building on this distinction, we propose \ours, a two-stage training framework that improves evidence reading through supervised fine-tuning (Read SFT) and downstream evidence utilization through Group Relative Policy Optimization (Use GRPO).
Experimental results demonstrate that our 8B model generalizes across three unseen multimodal memory systems, outperforming Qwen3-VL-32B and approaching GPT-5-mini.
\end{abstract}
\section{Introduction}
\label{sec:introduction}
Recent advances in vision-language models (VLMs) have expanded their applications beyond visual understanding, enabling VLM-based agents to perform interactive tasks such as computer use and robotic manipulation~\cite{cheng2024seeclick,qin2025ui,feng2025reflective}.
These tasks require agents to maintain relevant information across extended interactions, including visual observations and textual logs~\cite{li2024optimus,anwar2025remembr,zeng2026mementogui,liu2026goal2skill}.
To support such capabilities, multimodal memory has emerged as an important component of long-horizon agents, allowing them to preserve and revisit past experiences.
Memory construction methods organize relevant information from past experiences into structured and reusable representations~\cite{min2026reflectcap,liu2025memverse,wang2025mirix,liu2026omni}.
Complementing these efforts, memory retrieval methods focus on locating and selecting task-relevant information from stored memories~\cite{yeo2026universalrag,kang2026v,jeong2025videorag}.
Together, these efforts have made relevant information from past experiences more accessible to multimodal agents.

While advances in memory construction and retrieval have improved access to relevant multimodal evidence, their benefits depend on how effectively VLMs can use this evidence to answer downstream questions.
Indeed, prior studies have reported persistent challenges in answering questions using multimodal memory~\cite{liu2026worldmemarena,tang2026dmv,choi2026naive}.
These findings suggest that improving memory construction and retrieval alone may not be sufficient to improve downstream performance.
Therefore, improving memory utilization should be considered a complementary objective alongside memory construction and retrieval.

To investigate this challenge, we evaluate how well VLMs answer questions when necessary multimodal evidence is already provided.
Even with such evidence, GPT-5-mini achieves only 70.2\% accuracy, failing to answer a substantial portion of questions correctly.
Interestingly, replacing memory images with reference captions improves its accuracy to 91.6\%, suggesting that VLMs struggle to extract relevant information directly from visual representations.
However, such gains are difficult to achieve in practice: while reference captions preserve all visual details needed to answer the question, model-generated captions may omit essential information required for future questions, even yielding lower accuracy than the original images.
These observations suggest that converting images into text alone cannot reliably address the challenges of multimodal memory utilization.

These findings raise a more specific question about multimodal memory utilization: \textit{where do VLMs fail when sufficient multimodal evidence is already available?}
Building on prior findings that accessible knowledge is not always utilized in downstream tasks~\cite{kazemnejad2023measuring}, we examine two complementary aspects of memory utilization: reading answer-relevant evidence and using it to answer downstream questions.
Specifically, we introduce read probes to assess whether VLMs can directly recover answer-relevant information from the provided memory (Figure~\ref{fig:teaser}-(a)) and separately evaluate their ability to answer the original downstream questions.
As illustrated in Figure~\ref{fig:teaser}-(b), the baseline VLM fails to identify the relevant visual information and also answers the original question incorrectly, despite having access to the necessary evidence.
This example highlights the need to improve VLMs' ability to both accurately read relevant evidence from memory and use it for downstream question answering.

To strengthen both capabilities, we propose \ours(\textbf{Mem}ory \textbf{R}eading and \textbf{U}tilization), a two-stage framework for improving multimodal memory utilization.
First, we employ supervised fine-tuning on read probes, using answers derived from reference captions to teach VLMs to recover relevant information directly from multimodal memory (Read SFT).
Building on this, we then apply Group Relative Policy Optimization (GRPO)~\cite{shao2024deepseekmath} with question-answering rewards to encourage models to leverage the provided evidence when answering the question (Use GRPO).
Our results demonstrate the effectiveness of \ours in improving memory-based question answering: with the necessary evidence provided, our method improves Qwen3-VL-8B-Instruct by 11.8 percentage points over its base model.
Importantly, \ours also improves performance across three unseen multimodal memory systems, enabling our 8B model to outperform Qwen3-VL-32B-Instruct on average and achieve comparable performance to GPT-5-mini, as shown in Figure~\ref{fig:teaser}-(c).
Our analyses further provide insight into why this training sequence is effective: Read SFT improves evidence reading and expands correct-answer coverage under sampling, while Use GRPO further improves downstream answer correctness.
Together, these findings suggest that improving VLMs' ability to use multimodal memory can complement advances in memory construction and retrieval across different memory systems.
\section{Related Work}
\label{sec:related}
\paragraph{Multimodal memory for VLM agents.}
Recent studies have explored multimodal memory to support VLM agents in long-horizon tasks.
Memory construction methods organize experiences into structured and reusable representations~\cite{liu2025memverse,wang2025mirix,liu2026omni}.
Memory retrieval methods aim to retrieve task-relevant evidence from stored experiences~\cite{jain2025augustus,yeo2026universalrag, jeong2025videorag}.
Beyond construction and retrieval, recent studies have examined how visual memory representations affect downstream performance.
AGMem~\cite{choi2026naive} investigates the limitations of naively incorporating full-image memories into GUI agents, while DMV-Bench~\cite{tang2026dmv}, CapMem~\cite{liang2026capmem}, and SwitchMem~\cite{roh2026learning} explore trade-offs in visual and textual memory representations.
Building on these findings, our work complements existing memory-centric approaches by directly improving the ability of VLMs to utilize the provided multimodal evidence.

\paragraph{Learning to Utilize Multimodal Evidence.}
Memory-based question answering over long-horizon agent trajectories requires VLMs to use visual and textual evidence.
Prior work has shown that attending to visual evidence does not necessarily lead to correct predictions~\cite{liu2026seeing}.
ViGoRL~\cite{sarch2026grounded} and PEARL~\cite{zhang2026perceptual} improve visual grounding by incorporating perception-oriented verification in reinforcement learning, while other approaches optimize perception and reasoning separately~\cite {wu2026seeing,jiang2026videop2r,yoon2026pdcr}.
Complementary studies address visual forgetting through attention-aware optimization ~\cite{yuan2026laser}, visual evidence reinjection~\cite{pmlr-v267-zou25e}, latent memory mechanisms~\cite{yu2026vismem}, or reinforcement learning~\cite{chen2026remember}.
In contrast, our work focuses on training VLMs to read answer-relevant evidence from retrieved multimodal memories and use it for downstream question answering.
\section{Understanding Visual Memory Utilization}
\label{sec:setup}
\subsection{Problem Setup}
\label{sec:problem_setup}
We investigate how vision-language models (VLMs) utilize retrieved memories in multimodal agent settings.
Given an interaction history $H$ comprising visual observations and textual interactions, a multimodal memory system $\mathcal{M}$ retrieves relevant memory $M$ for a query $q$.
The memory system preserves both image-based and textual evidence from past interactions, allowing the retrieved memory to contain information from both modalities.
A downstream VLM $f_\theta$ then generates an answer $\hat{y}$ conditioned on $q$ and $M$:
\begin{equation}
    M = \mathcal{M}(H, q), \qquad
    \hat{y} = f_\theta(q, M).
\end{equation}
Our goal is to improve $f_\theta$'s ability to use the retrieved evidence effectively without modifying the underlying memory system $\mathcal{M}$.
\begin{figure}[t]
\centering
\includegraphics[width=\columnwidth]{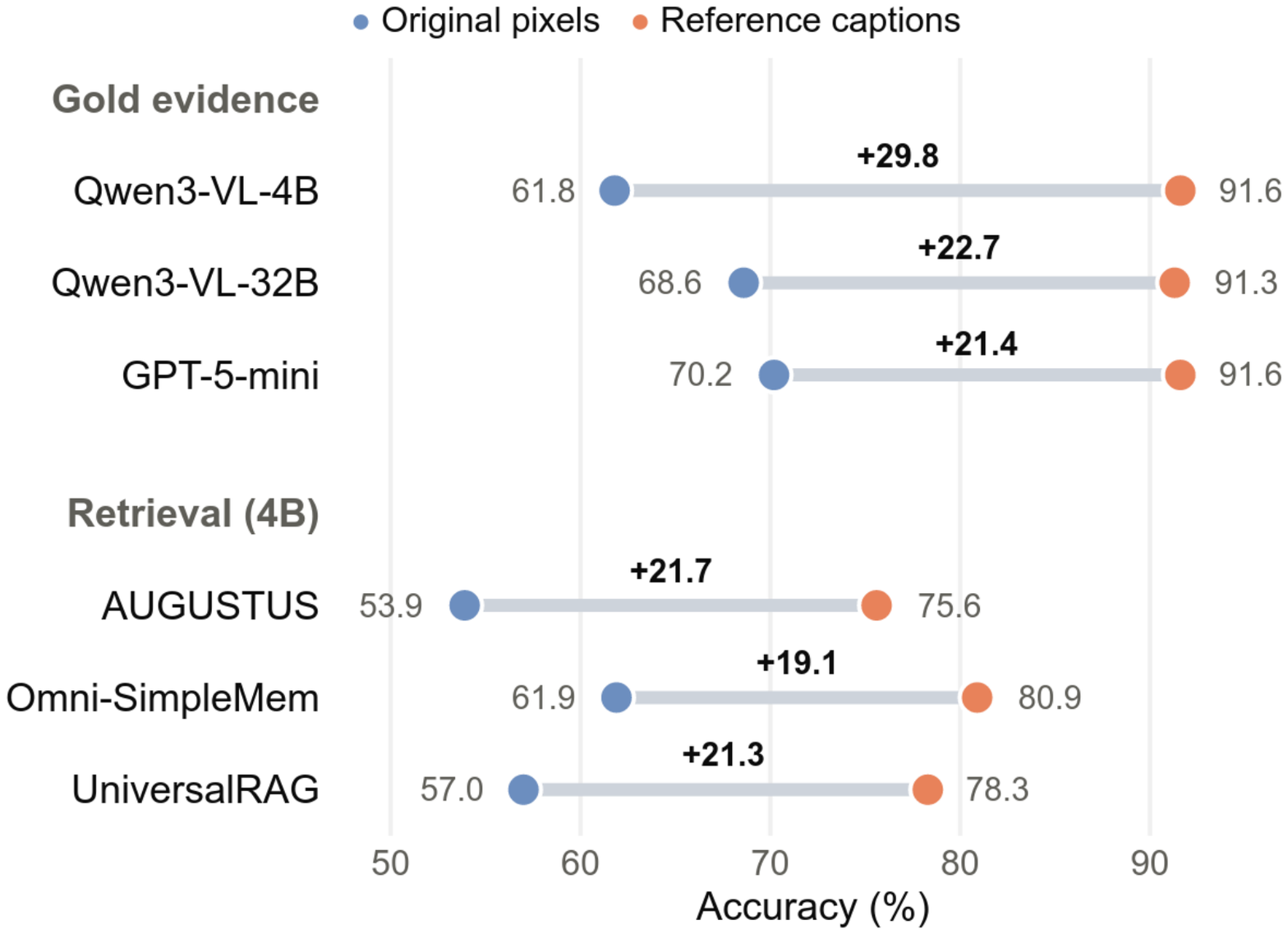}
\caption{
\textbf{Sufficiency does not guarantee successful memory utilization.}
The top group compares three VLMs under Gold Evidence, while the bottom group compares three multimodal memory systems using Qwen3-VL-4B-Instruct.
}
\label{fig:sufficiency_gap}
\vspace{-0.45cm}
\end{figure}
\subsection{Sufficiency Is Not Usability}
\label{sec:sufficiency}
We first investigate whether sufficient evidence in multimodal memory guarantees successful downstream question answering.
To control failures caused by insufficient evidence, we identify answer-sufficient memory instances in WorldMemArena (WMA)~\cite{liu2026worldmemarena}.
Starting from 580 pilot questions, we exclude 45 questions answered correctly without memory access.
For each remaining instance, GPT-6-Luna~\cite{openai2026gpt6luna} assesses whether the memory contains sufficient information to answer the question, using the gold answer and dataset-provided reference captions in place of visual observations.
We apply this assessment separately to WMA's Gold Evidence, which provides human-annotated oracle evidence, and three multimodal memory systems: AUGUSTUS~\cite{jain2025augustus}, Omni-SimpleMem~\cite{liu2026omni}, and UniversalRAG~\cite{yeo2026universalrag}.
All three systems use Qwen3-VL-4B-Instruct for memory construction to facilitate a controlled evaluation of memory utilization.
We retain only instances judged to contain answer-sufficient evidence for subsequent analysis; further details are provided in Appendix~\ref{app:memory_sufficiency}.

The downstream VLMs are then evaluated on these answer-sufficient instances using the original visual observations.
As shown in Figure~\ref{fig:sufficiency_gap}, accuracy remains at only 61.8--70.2\% across three VLMs in the Gold Evidence setting.
Similar limitations persist across the three memory systems, where Qwen3-VL-4B achieves only 53.9--61.9\% accuracy.
To further investigate this gap, the original visual observations are replaced with reference captions while keeping the question and remaining textual memory unchanged.
This replacement improves accuracy by over 20 percentage points for all three VLMs under Gold Evidence, with similar gains of approximately 19--22 percentage points across the three memory systems.
These findings reveal a substantial gap between the availability of sufficient visual evidence and VLMs' ability to effectively use it.

\begin{table}[t]
\centering
\renewcommand{\arraystretch}{1.12}
\resizebox{\columnwidth}{!}{
\begin{tabular}{@{}lccc@{}}
\toprule
\textbf{Memory Representation}
& \textbf{Sufficiency (\%)}
& \textbf{Acc (\%)}
& $\boldsymbol{\Delta}$ \textbf{(pp)} \\
\midrule
Original Pixels
& -- & 53.9 & -- \\
Reference Captions
& 100.0 & 75.6 & +21.7 \\
\midrule
\rowcolor{lightgray} \multicolumn{4}{@{}l}{\emph{Caption Replacement}} \\
Qwen3-VL-4B Captions
& 67.3 & 48.6 & \textcolor{red}{-5.3} \\
GPT-4o-mini Captions
& 60.8 & 46.9 & \textcolor{red}{-7.0} \\
\midrule
\rowcolor{lightgray} \multicolumn{4}{@{}l}{\emph{Caption Augmentation}} \\
Pixels + Qwen3-VL-4B Captions
& -- & 52.1 & \textcolor{red}{-1.8} \\
\bottomrule
\end{tabular}
}
\caption{
\textbf{Limitations of caption-based memory.}
Reference captions substantially improve downstream QA accuracy, whereas model-generated captions reduce answer sufficiency and accuracy.
}
\label{tab:captioning}
\vspace{-0.45cm}
\end{table}

\subsection{Limits of Captioning}   
The substantial gains from reference captions in Section~\ref{sec:sufficiency} suggest that converting visual observations into text could improve memory utilization.
We therefore investigate whether model-generated captions can provide similar benefits.
Specifically, using the AUGUSTUS memory system and Qwen3-VL-4B as the downstream VLM, we evaluate two strategies: \emph{caption replacement}, where captions are generated by Qwen3-VL-4B or an external proprietary model (GPT-4o-mini), and \emph{caption augmentation}, which adds Qwen3-VL-4B captions alongside the original images.
For caption replacement, we repeat the sufficiency assessment from Section~\ref{sec:sufficiency} on memories reconstructed with model-generated captions, measuring the proportion of answer-sufficient instances.
As shown in Table~\ref{tab:captioning}, only 60.8--67.3\% of these instances remain answer-sufficient after replacement, while downstream accuracy decreases by 5.3--7.0 percentage points compared to the original visual memories.
Moreover, caption augmentation achieves 52.1\% accuracy, providing no improvement over the original visual memory.
This contrast reveals a practical challenge: unlike reference captions, model-generated captions may omit or misrepresent visual details essential for downstream question answering.
In practice, captions must be generated without knowing future questions, making it difficult to ensure that all potentially relevant information is preserved.
These results suggest that converting images into text alone cannot reliably address the challenges of multimodal memory utilization.
Additional details on caption-based memories are provided in Appendix~\ref{app:captioning}.

\subsection{Where Does Memory Use Fail?}
\label{sec:failure_diagnosis}
To better understand why VLMs fail despite having sufficient evidence in multimodal memory, we investigate how models utilize the provided memory.
Inspired by prior findings that models may fail to use knowledge they can access~\cite{kazemnejad2023measuring}, we distinguish between reading answer-relevant evidence from memory and using it to answer the original question.
For each answer-sufficient instance, GPT-6-Luna extracts answer-critical facts from the reference-caption memory and generates a short question for each fact, termed a \emph{read probe}, to assess evidence reading.
Using the original multimodal memory, we assess \textit{Read} by whether the VLM correctly answers all read probes and \textit{Use} by whether it correctly answers the original question.
Based on these outcomes, we categorize instances into four groups: \emph{Read Failure} (0, 0), \emph{Use Failure} (1,0), \emph{QA Success (Read \ding{55})} (0, 1), and \emph{QA Success (Read \ding{51})} (1,1), each pair denoting (Read, Use).
To validate this diagnosis, we conduct a human audit, finding high judge agreement (95.1\%) and read-probe reference correctness (92.5\%). 
Details of read-probe construction and the human audit are provided in Appendix~\ref{app:read_use_diagnosis}.
Figure~\ref{fig:failure_diagnosis} shows the resulting distribution for Qwen3-VL-4B across WMA's Gold Evidence and three multimodal memory systems.
\emph{Read Failure} accounts for 20.9--21.9\% of instances, while \emph{Use Failure} accounts for 28.9--32.2\%, with similar distributions across all four memory settings.
Although some instances achieve correct downstream answers despite failing to read (8.1--10.2\%), these cases constitute a relatively small proportion compared to the two failure categories.
Together, these results highlight the need to improve both evidence reading and downstream memory utilization.

\begin{figure}[t]
\centering
\includegraphics[width=\columnwidth]{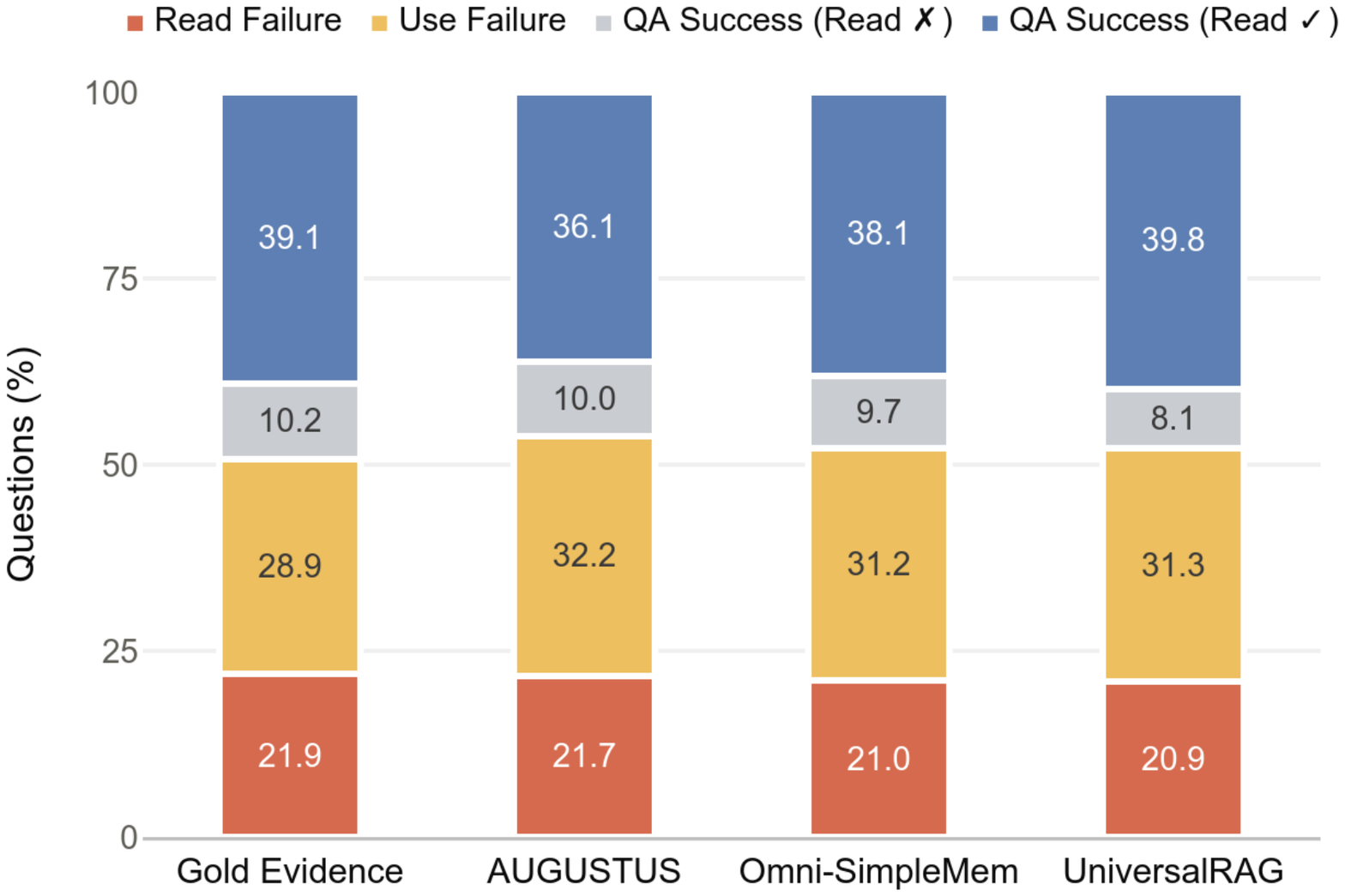}
\caption{
\textbf{Failure analysis in multimodal memory utilization.}
Both Read Failure and Use Failure account for substantial portions of instances, highlighting the need to improve both capabilities. Details about each category are provided in Appendix~\ref{app:failure_categorization}
}
\label{fig:failure_diagnosis}
\vspace{-0.45cm}
\end{figure}

\section{MemRU}
\label{sec:method}
\begin{figure*}[ht]
\centering
\includegraphics[width=0.95\linewidth]{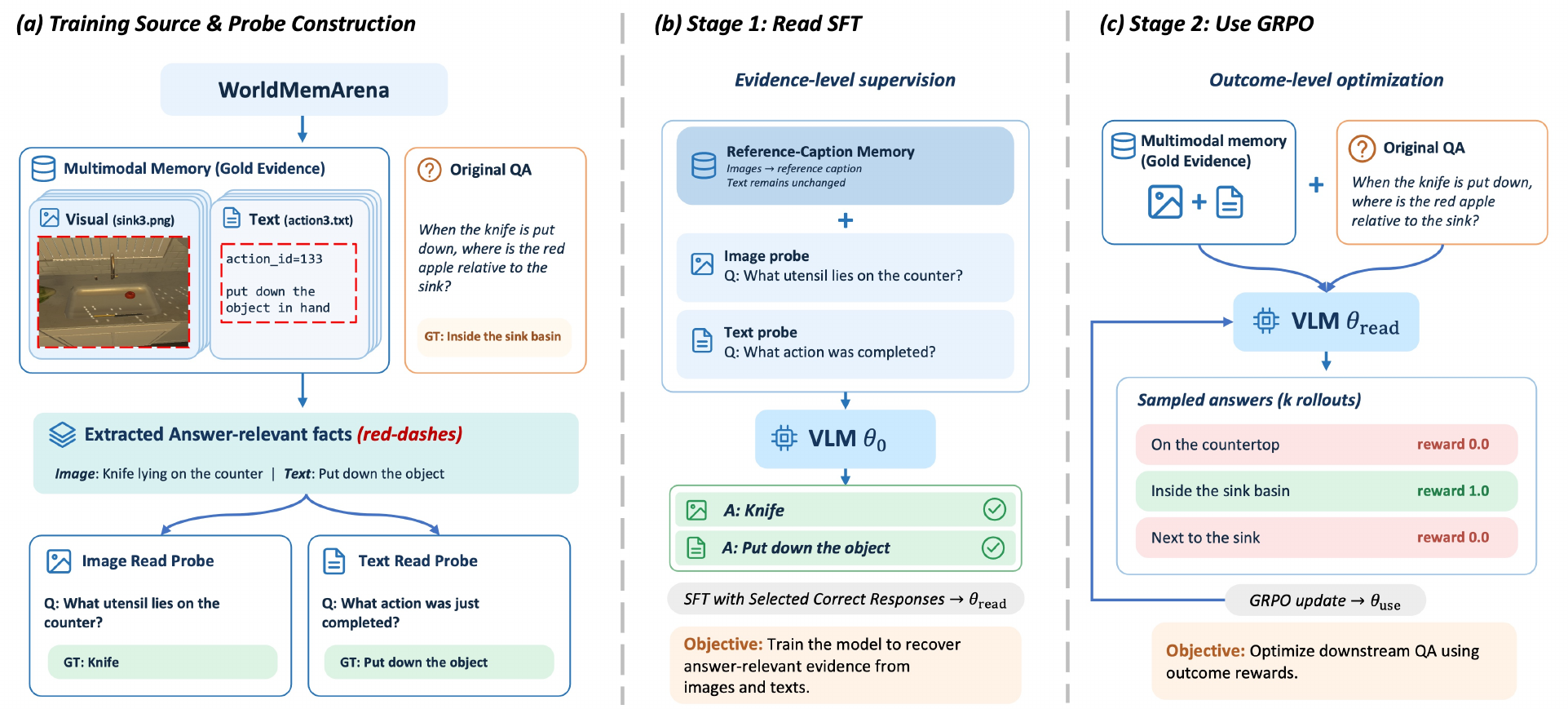}
\caption{
\textbf{Overview of \ours.}
The framework consists of two stages: Read SFT learns to extract relevant evidence through image and text read probes, while Use GRPO optimizes downstream QA using outcome-based rewards.
Red dashed boxes highlight the answer-relevant evidence used to construct read probes.
}
\label{fig:overview}
\vspace{-0.45cm}
\end{figure*}

Motivated by the analysis in Section~\ref{sec:setup}, we propose \ours, a two-stage training framework that improves how VLMs read and use retrieved multimodal memory.
As illustrated in Figure~\ref{fig:overview}, the first stage employs supervised fine-tuning (Read SFT) to improve evidence reading, while the second stage uses reinforcement learning (Use GRPO) to enhance downstream evidence utilization.
Rather than altering the structure or representation of memory, \ours focuses on improving the VLM's ability to utilize retrieved evidence across different memory systems.
\subsection{Learning to Read Memory}
\label{sec:read_sft}
As shown in Section~\ref{sec:failure_diagnosis}, VLMs often fail to accurately read relevant evidence from retrieved multimodal memory, even when sufficient information is available.
To address this limitation, we introduce Read SFT, which provides direct supervision for reading answer-relevant evidence from multimodal memory.
We construct a separate training set from the training partition of WMA, using Gold Evidence as the multimodal memory $M_i$ and generating image and text read probes $\mathcal{P}_i = \{(p_{ij}, a_{ij})\}_{j=1}^{J_i}$ following Section~\ref{sec:failure_diagnosis}.
Since downstream questions may require evidence from both visual observations and textual interactions, Read SFT incorporates both image and text read probes. 
This provides supervision across heterogeneous memory content rather than focusing solely on visual evidence.
For each read probe $p_{ij}$ with reference answer $a_{ij}$, we first obtain a response from the original VLM using $M_i^{\mathrm{txt}}$, where only visual observations are replaced with reference captions.
We retain responses that correctly answer their corresponding probes:
\begin{equation}
\begin{aligned}
\hat{a}_{ij} &= f_{\theta_0}(p_{ij}, M_i^{\mathrm{txt}}), \\
\mathcal{S}_i &=
\{(p_{ij},\hat{a}_{ij}) \mid
\mathrm{Correct}(\hat{a}_{ij},a_{ij})=1\},
\end{aligned}
\end{equation}
where $\theta_0$ denotes the parameters of original VLM, and $\mathrm{Correct}(\cdot,\cdot)$ evaluates response correctness.
This selection provides supervision from responses the model can already generate correctly using reference-caption memory, allowing Read SFT to learn to recover the corresponding evidence from the original multimodal memory.

The Read SFT dataset is then constructed as $\mathcal{D}_{\mathrm{read}}
=
\bigcup_i
\{(M_i,p,\hat{a})\mid(p,\hat{a})\in\mathcal{S}_i\}$ by pairing the selected responses with the original multimodal memory.
Importantly, reference captions are used during probe construction and response selection, whereas Read SFT operates on the original multimodal memory.
We fine-tune the VLM $\theta_0$ on $\mathcal{D}_{\mathrm{read}}$ using the standard supervised fine-tuning objective.
The resulting parameters, denoted by $\theta_{\mathrm{read}}$, initialize the subsequent Use GRPO stage.
Further details of Read SFT data construction are provided in Appendix~\ref{app:read_sft}.

\subsection{Learning to Use Memory}
\label{sec:use_grpo}
While Read SFT improves evidence reading, successfully reading relevant evidence does not necessarily lead to correct downstream answers.
To bridge this gap between reading and utilizing evidence, the second stage employs Group Relative Policy Optimization (GRPO)~\cite{shao2024deepseekmath}, termed \emph{Use GRPO}, to optimize the model $\theta_{\mathrm{read}}$ using outcome rewards to reinforce successful responses to the original questions.

Using the original questions from the training partition of WMA, with Gold Evidence as the multimodal memory $M_i$, the model generates $K$ candidate responses $\{\hat{y}_i^{(k)}\}_{k=1}^{K}$ to the original question $q_i$.
Each response is assigned a reward based on answer correctness, with a penalty for excessive response length:
\begin{equation}
r_i^{(k)}
=
s(q_i,y_i,\hat{y}_i^{(k)})
+
\ell(\hat{y}_i^{(k)},y_i),
\end{equation}
where $y_i$ is the gold answer, $s(\cdot)$ is an LLM-based correctness score ranging from $0$ to $1$, and $\ell(\cdot)$ is a length penalty inspired by DAPO~\cite{yu2026dapo}.
The length penalty increases linearly when the response exceeds twice the gold-answer length, with a maximum deduction of $0.5$.
Further details on reward computation are provided in Appendix~\ref{app:use_grpo}.

Based on these rewards, GRPO computes relative advantages among the candidate responses to optimize the VLM for downstream question answering.
The trained model, denoted by $f_{\theta_{\mathrm{use}}}$, is used to answer downstream questions directly from retrieved multimodal memory.

\begin{table*}[t]
\centering
\scriptsize
\renewcommand{\arraystretch}{1.12}
\resizebox{\textwidth}{!}{
\begin{tabular}{lccccc}
\toprule
\multirow{3}{*}{\textbf{Model / Method}}
& \multirow{3}{*}{\textbf{Gold Evidence}}
& \multicolumn{4}{c}{\textbf{Retrieved Memory Systems}} \\
\cmidrule(lr){3-6}
& & \textbf{AUGUSTUS}~\cite{jain2025augustus}
& \textbf{Omni-SimpleMem}~\cite{liu2026omni}
& \textbf{UniversalRAG}~\cite{yeo2026universalrag}
& \textbf{Cross-system Avg.} \\
\midrule

\multicolumn{6}{l}{\textit{General-purpose VLMs}} \\
GPT-5-mini
& 73.3 & 61.1 & 62.5 & 63.7 & 62.4 \\
Qwen3-VL-32B-Instruct
& 71.4 & 58.0 & 58.9 & 61.2 & 59.4 \\

\midrule
\multicolumn{6}{l}{\textit{Qwen3-VL-4B-Instruct}} \\
Base
& 62.9 & 52.1 & 52.4 & 53.5 & 52.7 \\
Use SFT
& 64.1 & 54.8 & 55.3 & 56.1 & 55.4 \\
Use GRPO
& 69.6 & 55.6 & 57.1 & 55.3 & 56.0 \\
Use SFT $\rightarrow$ Use GRPO
& 72.5 & \textbf{56.7} & 58.1 & 57.0 & 57.3 \\
OPD-V (4B)
& 62.9 & 52.5 & 54.8 & 54.1 & 53.8 \\
MultiHopSpatial (4B)
& 64.1 & 52.7 & 55.2 & 53.4 & 53.8 \\
\rowcolor{gray!10}
\textbf{\ours}
& \textbf{73.3} & 56.4 & \textbf{60.0}
& \textbf{58.8} & \textbf{58.4} \\

\midrule
\multicolumn{6}{l}{\textit{Qwen3-VL-8B-Instruct}} \\
Base
& 62.4 & 52.4 & 54.6 & 54.5 & 53.8 \\
Use SFT
& 66.8 & 54.1 & 55.8 & 55.2 & 55.1 \\
Use GRPO
& 73.7 & 57.1 & 57.3 & 57.9 & 57.4 \\
Use SFT $\rightarrow$ Use GRPO
& 73.4 & 58.0 & 58.2 & 59.9 & 58.7 \\
OPD-V (8B)
& 64.5 & 53.9 & 54.8 & 56.1 & 54.9 \\
MultiHopSpatial (8B)
& 66.9 & 55.4 & 57.4 & 57.9 & 56.9 \\
\rowcolor{gray!10}
\textbf{\ours}
& \textbf{74.2} & \textbf{60.8} & \textbf{59.9}
& \textbf{60.8} & \textbf{60.5} \\
\bottomrule
\end{tabular}
}
\caption{
\textbf{Main results on WorldMemArena (WMA).}
QA accuracy (\%) is reported under Gold Evidence and three memory systems.
Cross-system Avg. denotes the mean accuracy across the three retrieved memory systems, excluding Gold Evidence.
Bold indicates the best performance within each 4B or 8B group. 
We report 95\% CIs for individual accuracies and paired accuracy differences in Appendix~\ref{app:95_ci}.
}
\label{tab:main_results}
\vspace{-0.45cm}
\end{table*}

\section{Experiments}
\label{sec:experiments}

\subsection{Experimental Setup}
\noindent\textbf{Models and Datasets.}
We conduct experiments on WorldMemArena (WMA)~\cite{liu2026worldmemarena}, spanning GUI, embodied, and lifelong scenarios.
To prevent data leakage, we split WMA at the scenario level into training, validation, and test sets at a 70:10:20 ratio, yielding 2,073 test questions after filtering.
Unlike the diagnostic pilot in Section \ref{sec:setup}, the main evaluation uses the same 2,073 test questions across all memory systems without filtering by answer sufficiency.
We train \ours on Qwen3-VL-4B-Instruct and Qwen3-VL-8B-Instruct~\cite{bai2025qwen3}.
\emph{Read SFT} uses balanced image- and text-based probe examples, while \emph{Use GRPO} trains on 6,722 original questions containing visual evidence.

\noindent\textbf{Memory Systems.}
Both training stages use WMA's Gold Evidence, which provides question-relevant oracle evidence, avoiding adaptation to any particular retrieval pipeline.
To test whether the learned reading and utilization capabilities transfer across memory systems, we evaluate on retrieved memories from AUGUSTUS~\cite{jain2025augustus}, Omni-SimpleMem~\cite{liu2026omni}, and UniversalRAG~\cite{yeo2026universalrag}.
These memories are preconstructed using Qwen3-VL-4B-Instruct and held fixed across evaluated VLMs to control for differences in memory construction.

\noindent\textbf{Baselines and Implementation.}
We compare against four controlled training baselines: the original model (Base), QA-supervised fine-tuning (Use SFT), direct reinforcement learning (Use GRPO), and Use SFT followed by Use GRPO.
We also evaluate vision-focused post-trained VLMs: OPD-V~\cite{bi2026opd}, which improves visual reasoning through on-policy self-distillation, and MultiHopSpatial~\cite{lee2026multihopspatial}, which uses GRPO for spatial reasoning.
We include stronger general-purpose VLMs, GPT-5-mini~\cite{singh2025openai} and Qwen3-VL-32B-Instruct~\cite{bai2025qwen3}.
Both stages of \ours employ LoRA~\cite{hu2022lora}, freezing the vision encoder and multimodal merger to focus on utilizing existing visual representations.
Read SFT is trained for two epochs, followed by one epoch of Use GRPO; SFT and GRPO variants share their respective hyperparameters, with GRPO checkpoints selected independently by validation performance.
Additional experimental details are provided in Appendix~\ref{app:experimental_details}.

\subsection{Main Results}
\label{sec:main_results}
Table~\ref{tab:main_results} presents the QA accuracy of \ours and baseline methods across four multimodal memory settings.
Under Gold Evidence, \ours consistently outperforms all corresponding training baselines.
Notably, the 4B model matches GPT-5-mini, while the 8B model surpasses both GPT-5-mini and Qwen3-VL-32B-Instruct.
Beyond Gold Evidence, these performance gains also generalize to unseen multimodal memory systems.
Across AUGUSTUS, Omni-SimpleMem, and UniversalRAG, \ours achieves the highest average accuracy among all 4B and 8B baselines, outperforming both controlled training baselines and existing post-trained VLMs.
In particular, the 8B model achieves 60.5\% cross-system average accuracy, surpassing Qwen3-VL-32B-Instruct while remaining competitive with GPT-5-mini, with a gap of 1.9\%.
Compared with the Use SFT $\rightarrow$ Use GRPO baseline, \ours achieves higher cross-system average accuracy at both scales, with a statistically supported improvement (Appendix~\ref{app:95_ci}).
\begin{figure*}[t]
\centering
\includegraphics[width=0.85\linewidth]{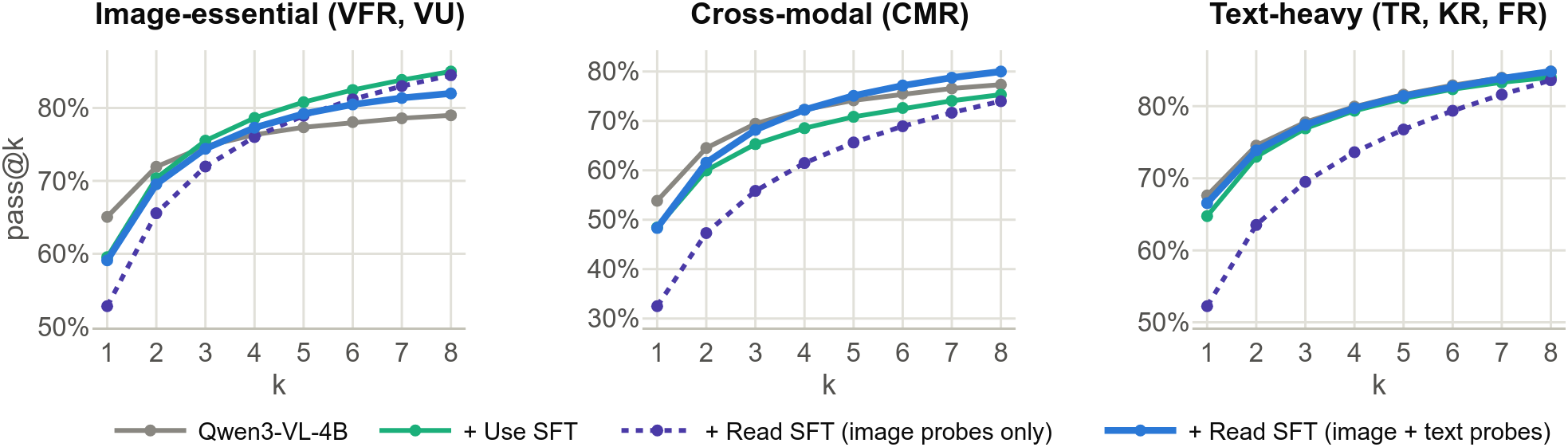}
\caption{
\textbf{Effect of read supervision on downstream answer coverage.}
Pass@$k$ of Qwen3-VL-4B-Instruct across image-essential (VFR, VU), cross-modal (CMR), and text-heavy (TR, KR, FR) questions.
Joint Read SFT achieves higher answer coverage at larger sampling budgets, particularly on cross-modal and text-heavy questions.
}
\label{fig:passk}
\vspace{-0.45cm}
\end{figure*}
\subsection{Why Does \ours Work?}
\subsubsection{Effectiveness of Read Supervision}
\begin{table}[t]
\centering
\footnotesize
\begin{tabular}{clccc}
\toprule
\textbf{Model} & \textbf{Method}
& \textbf{Image} & \textbf{Text} & \textbf{Total} \\
\midrule

\multirow{7}{*}{\textbf{4B}}
& Base & 60.9 & 94.1 & 81.1 \\
& \hspace{1.0em}+ GRPO & 63.2 & 94.4 & 82.2 \\
\cmidrule{2-5}
& Use SFT & 64.0 & 94.2 & 82.4 \\
& \hspace{1.0em}+ GRPO & 63.9 & 95.0 & 83.2 \\
\cmidrule{2-5}
& Read SFT & 67.9 & 96.0 & 84.9 \\
\rowcolor{gray!10}
& \hspace{1.0em}+ GRPO
& \textbf{69.8} & \textbf{96.3} & \textbf{85.9} \\

\midrule

\multirow{7}{*}{\textbf{8B}}
& Base & 60.1 & 94.0 & 80.7 \\
& \hspace{1.0em}+ GRPO & 66.1 & 95.7 & 84.1 \\
\cmidrule{2-5}
& Use SFT & 64.2 & 94.3 & 82.5 \\
& \hspace{1.0em}+ GRPO & 66.2 & 94.8 & 83.6 \\
\cmidrule{2-5}
& Read SFT & 67.4 & 95.8 & 84.6 \\
\rowcolor{gray!10}
& \hspace{1.0em}+ GRPO
& \textbf{69.8} & \textbf{96.1} & \textbf{85.8} \\

\bottomrule
\end{tabular}
\caption{
\textbf{Effectiveness of read supervision on memory reading.}
Reading accuracy (\%) on image and text probes across different training configurations.
}
\label{tab:read_supervision}
\vspace{-0.45cm}
\end{table}

Our failure diagnosis revealed that VLMs often fail to read answer-relevant evidence even when sufficient information is available in memory.
To investigate whether explicit read supervision addresses this bottleneck, we evaluate the base model, Use SFT, and Read SFT, each with and without subsequent Use GRPO, using image and text read probes.
As shown in Table~\ref{tab:read_supervision}, Read SFT consistently achieves higher read probe accuracy than the alternatives at both model scales, with larger gains on visual probes.
Notably, Use GRPO further improves read probe accuracy in most settings, despite receiving no direct supervision from read probes.
These results demonstrate that explicit read supervision effectively strengthens the model's ability to accurately read both visual and textual evidence in memory.

\subsubsection{From Reading to Answer Coverage}
Improved evidence reading does not necessarily translate into higher downstream QA accuracy.
We therefore examine whether Read SFT increases the availability of correct downstream answers under sampling, measured by pass@k.
The comparison includes the base model, Use SFT, and two Read SFT variants trained on image-only or joint image-text read probes, reporting results by the evidence modalities required to answer each question.
As shown in Figure~\ref{fig:passk}, joint image-text read supervision achieves higher answer coverage than Use SFT at larger sampling budgets, particularly on cross-modal tasks.
In contrast, Use SFT and image-only Read SFT remain competitive on image-essential tasks but provide less answer coverage than joint Read SFT on tasks requiring textual evidence.
These findings suggest that Read SFT can make correct downstream responses more accessible under sampling, even without consistently improving pass@1, providing a useful initialization for subsequent outcome-based optimization.



\begin{table}[t]
\centering
\scriptsize
\setlength{\tabcolsep}{6pt}
\renewcommand{\arraystretch}{1.12}
\resizebox{\columnwidth}{!}{
\begin{tabular}{clccc}
\toprule
\textbf{Model} & \textbf{Stage 2 Training}
& \textbf{Read} & \textbf{Use}
& \textbf{Use Given Read} \\
\midrule

\multirow{3}{*}{\textbf{4B}}
& None
& 68.6 & 64.8 & 72.3 \\
& Use SFT
& 67.0 & 64.0 & 71.2 \\
\rowcolor{gray!10}
& \textbf{Use GRPO}
& \textbf{69.8} & \textbf{73.1} & \textbf{80.2} \\

\midrule

\multirow{3}{*}{\textbf{8B}}
& None
& 67.7 & 60.6 & 67.9 \\
& Use SFT
& 66.5 & 66.1 & 72.5 \\
\rowcolor{gray!10}
& \textbf{Use GRPO}
& \textbf{69.7} & \textbf{74.2} & \textbf{79.8} \\

\bottomrule
\end{tabular}
}
\caption{
\textbf{Effect of stage 2 configuration on memory utilization.}
Comparison of reading, QA, and conditional QA accuracy (\%) across training strategies following Read SFT.
}
\label{tab:outcome_rl}
\vspace{-0.45cm}
\end{table}

\begin{table*}[t]
\centering
\scriptsize
\renewcommand{\arraystretch}{1.1}
\resizebox{\textwidth}{!}{
\begin{tabular}{
cl
cccc>{\columncolor{gray!10}}c
ccccccc>{\columncolor{gray!10}}c
}
\toprule
\multirow{2}{*}{\textbf{Model}}
& \multirow{2}{*}{\textbf{Method}}
& \multicolumn{5}{c}{\textbf{MemLens}~\cite{ren2026memlens}}
& \multicolumn{8}{c}{\textbf{OpenEQA}~\cite{majumdar2024openeqa}} \\
\cmidrule(lr){3-7}
\cmidrule(lr){8-15}
&
& \textbf{IE} & \textbf{KU} & \textbf{MSR}
& \textbf{TR} & \textbf{Overall}
& \textbf{AR} & \textbf{Func.} & \textbf{Loc.}
& \textbf{OR} & \textbf{OSR} & \textbf{Spatial}
& \textbf{WK} & \textbf{Overall} \\
\midrule


\multirow{6}{*}{\textbf{4B}}
& Base
& 61.0 & 50.0 & 20.3 & 51.0 & 48.6
& 72.9 & 45.2 & 59.3 & 51.9 & 70.6 & 42.7 & 48.4 & 56.5 \\

& Use GRPO
& 65.9 & 55.2 & \textbf{39.9} & 56.7 & 56.2
& 72.5 & 47.9 & 62.0 & 55.0 & 69.0 & 42.3 & 51.6 & 57.8 \\

& Use SFT $\rightarrow$ Use GRPO
& 58.1 & \textbf{59.5} & 34.3 & 67.0 & 55.9
& 70.0 & 46.5 & 61.2 & 52.8 & 66.7 & 40.5 & \textbf{53.1} & 56.4 \\

& OPD-V
& \textbf{66.7} & 50.0 & 25.2 & 59.3 & 53.4
& \textbf{75.4} & 47.9 & 57.4 & \textbf{57.1}
& 73.8 & 41.4 & 51.2 & 58.3 \\

& MultiHopSpatial
& 64.6 & 56.0 & 26.6 & 63.4 & 55.1
& 73.8 & 46.1 & 58.9 & 52.4 & 72.6 & 42.3 & 49.8 & 57.2 \\

\cmidrule(lr){3-7}
\cmidrule(lr){8-15}
& \textbf{\ours}
& 57.7 & 52.6 & 38.5 & \textbf{71.6} & \textbf{56.8}
& 72.1 & \textbf{48.4} & \textbf{64.6} & 56.7
& \textbf{75.0} & \textbf{44.1} & 52.1 & \textbf{59.2} \\

\midrule


\multirow{6}{*}{\textbf{8B}}
& Base
& 61.0 & 60.3 & 33.6 & 54.6 & 53.5
& 69.6 & 44.7 & 59.3 & 55.0 & 74.6 & 40.9 & 54.0 & 57.5 \\

& Use GRPO
& 59.8 & 59.5 & 35.0 & 67.5 & 56.8
& 71.7 & 51.2 & \textbf{67.3} & 58.4
& 67.1 & \textbf{46.8} & 56.8 & 60.4 \\

& Use SFT $\rightarrow$ Use GRPO
& 58.9 & \textbf{66.4} & 33.6 & 64.9 & 58.4
& 68.8 & 48.8 & 53.2 & 53.7 & 75.0 & 41.4 & 54.9 & 57.0 \\

& OPD-V
& \textbf{72.0} & 52.6 & 28.0 & 63.4 & 57.4
& \textbf{75.8} & 48.4 & 60.1 & \textbf{59.3}
& 78.2 & 45.0 & 54.5 & \textbf{60.8} \\

& MultiHopSpatial
& 69.1 & 52.6 & 29.4 & 59.8 & 55.7
& 72.5 & 47.0 & 54.4 & 58.4 & \textbf{79.4}
& 43.2 & 51.6 & 58.6 \\

\cmidrule(lr){3-7}
\cmidrule(lr){8-15}
& \textbf{\ours}
& 60.6 & 57.8 & \textbf{37.8} & \textbf{73.7}
& \textbf{59.1}
& 72.5 & \textbf{54.4} & 63.1 & 50.6
& 74.6 & 43.6 & \textbf{60.6} & 59.5 \\

\bottomrule
\end{tabular}
}
\caption{
\textbf{Cross-benchmark generalization on MemLens and OpenEQA.}
Category-wise QA accuracy (\%) of 4B and 8B models using oracle evidence for MemLens and 50 frames for OpenEQA.
Bold indicates the best performance within each model scale and category.
}
\label{tab:ood_generalization}
\vspace{-0.45cm}
\end{table*}

\subsubsection{Role of Outcome-based Reinforcement Learning}
While Read SFT can increase answer coverage under sampling, it does not directly optimize correctness on the original downstream questions.
To identify an effective training strategy for this purpose, we compare three Stage 2 configurations following Read SFT in Stage 1: no additional training, Use SFT, and Use GRPO.
We evaluate Read and Use accuracy, along with Use Given Read, defined as $P(\mathrm{Use}=1 \mid \mathrm{Read}=1)$, which measures QA accuracy conditioned on successful memory reading.
As shown in Table~\ref{tab:outcome_rl}, Use GRPO improves both Read and Use accuracy, while also increasing Use Given Read.
This suggests that the benefits of Use GRPO extend beyond overall QA accuracy, with improved downstream correctness among questions for which the model successfully reads the required evidence.
Together, these findings highlight the complementary roles of Read SFT and Use GRPO in improving evidence reading and downstream answer correctness conditioned on successful reading.

\subsection{Category-wise Performance Analysis}
To examine whether the improvements of \ours are concentrated in particular question types, we evaluate QA accuracy across six categories: Visual Factual Recall (VFR), Visual Update (VU), Cross-modal Reasoning (CMR), Temporal Reasoning (TR), Knowledge Reasoning (KR), and Factual Reasoning (FR).
We compare \ours against controlled training baselines, Qwen3-VL-32B-Instruct, and MultiHopSpatial, averaging results across AUGUSTUS, Omni-SimpleMem, and UniversalRAG.
As shown in Table~\ref{tab:evidence_modalities}, \ours achieves higher accuracy than the strongest controlled training baseline in all six categories when averaged across the three memory systems, demonstrating improvements across diverse memory reasoning tasks.
Furthermore, \ours achieves the highest accuracy on VU, TR, and KR among all evaluated models, surpassing Qwen3-VL-32B-Instruct and MultiHopSpatial in overall accuracy.
Overall, these findings suggest that the benefits of \ours extend across diverse memory reasoning categories, rather than being confined to a particular question type.
Detailed results for individual memory systems are provided in Appendix~\ref{sec:detailed_memory_systems}, showing that these improvements broadly extend across different memory systems.

\begin{table}[t]
\centering
\small
\setlength{\tabcolsep}{3pt}
\renewcommand{\arraystretch}{1.12}
\resizebox{\columnwidth}{!}{
\begin{tabular}{lccccccc}
\toprule
\textbf{Method}
& \textbf{\textit{VFR}} & \textbf{\textit{VU}} & \textbf{\textit{CMR}}
& \textbf{\textit{TR}} & \textbf{\textit{KR}} & \textbf{\textit{FR}}
& \textbf{Overall} \\
\midrule

Base (8B)
& 54.9 & 55.4 & 44.8 & 26.4 & 48.0 & 79.3 & 53.8 \\

Use GRPO (8B)
& 63.2 & 58.0 & 48.4 & 27.7 & 63.8 & 81.8 & 57.4 \\

Use SFT $\rightarrow$ GRPO (8B)
& 60.7 & 64.3 & 50.8 & 26.6 & 66.4 & 79.0 & 58.7 \\

\midrule

Qwen3-VL-32B
& \textbf{64.5} & 61.0 & \textbf{52.4}
& 28.5 & 65.4 & \textbf{83.1} & 59.4 \\

MultiHopSpatial (8B)
& 61.6 & 57.8 & 51.4 & 26.2 & 62.5 & 80.7 & 56.9 \\

\midrule

\rowcolor{gray!10}
\textbf{\ours{} (8B)}
& 63.4 & \textbf{64.7} & 51.7
& \textbf{29.2} & \textbf{69.8} & 82.8
& \textbf{60.5} \\

\bottomrule
\end{tabular}
}
\caption{
\textbf{Performance across each category.}
QA accuracy across six reasoning categories, averaged over AUGUSTUS, Omni-SimpleMem, and UniversalRAG.
Bold indicates the best result.
}
\label{tab:evidence_modalities}
\vspace{-0.45cm}
\end{table}

\subsection{Out-of-Distribution Generalization}
To examine whether the benefits of \ours transfer beyond the training benchmark, we evaluate its performance on two unseen benchmarks, MemLens~\cite{ren2026memlens} and OpenEQA~\cite{majumdar2024openeqa}.
We use oracle evidence for MemLens to assess downstream evidence utilization and follow the standard 50-frame evaluation setting for OpenEQA.
We examine category-wise performance across different memory utilization needs, including Information Extraction (IE), Knowledge Update (KU), Multi-session Reasoning (MSR), and Temporal Reasoning (TR) in MemLens.
OpenEQA further evaluates visual understanding and reasoning through Attribute Recognition (AR), Object Recognition (OR), Object State Recognition (OSR), Object Localization (Loc.), Functional Reasoning (Func.), Spatial Understanding (Spatial), and World Knowledge (WK).
We compare \ours against the base model, two GRPO-based training variants, and existing post-trained VLMs.

As shown in Table~\ref{tab:ood_generalization}, \ours generalizes across both benchmarks and model scales, achieving the highest overall accuracy in three of the four evaluation settings.
On MemLens, \ours shows particularly strong gains in temporal reasoning and multi-session reasoning.
On OpenEQA, \ours achieves the highest functional reasoning accuracy at both model scales.
In contrast, improvements are less consistent on information extraction and knowledge updating in MemLens, as well as object recognition and localization in OpenEQA.
This pattern aligns with our training objectives: Read SFT improves evidence reading, while Use GRPO encourages reasoning over multiple pieces of evidence. 
However, neither stage explicitly targets conflicting memory updates or fine-grained object grounding, limiting transfer to these tasks.
Together, these results suggest that the benefits of memory-oriented post-training extend beyond WMA-specific question patterns, with particularly strong transfer to temporally structured reasoning tasks.
\section{Conclusion}
\label{sec:conclusion}
Our study reveals a gap between sufficient evidence in multimodal memory and its effective utilization by VLMs.
Importantly, our analysis further shows that optimizing downstream QA with GRPO alone does not consistently address both evidence-reading and evidence-utilization failures.
Motivated by this distinction, we introduced \ours, which combines explicit read supervision with downstream outcome optimization to address these complementary challenges.
The improvements across unseen memory systems and benchmarks further demonstrate the potential of training VLMs to better utilize multimodal evidence.
Together, our findings suggest that advancing multimodal memory requires not only better memory construction and retrieval, but also a deeper understanding of how models read and use available evidence to produce correct downstream answers.

\section*{LLM Usage Statement}
We used ChatGPT (OpenAI) and Claude (Anthropic) to provide feedback on manuscript drafts and to assist with language editing and figure refinement. 
Claude Code (Anthropic) was additionally used to assist with experimental implementation, including code generation, debugging, and experiment execution.
The authors reviewed all AI-assisted outputs and take full responsibility for the research design, experimental results, interpretations, and final manuscript content.
{
    \small
    \bibliographystyle{ieeenat_fullname}
    \bibliography{main}
}

\clearpage
\setcounter{page}{1}
\maketitlesupplementary

\noindent\textbf{Contents}
\begin{itemize}[leftmargin=*,nosep]
    \item \hyperref[app:memory_sufficiency]{\ref*{app:memory_sufficiency}. Answer Sufficiency Assessment}
    \item \hyperref[app:captioning]{\ref*{app:captioning}. Caption-Based Memory Construction}
    \item \hyperref[app:read_use_diagnosis]{\ref*{app:read_use_diagnosis}. Read Probe Construction}
    \item \hyperref[app:read_sft]{\ref*{app:read_sft}. Read SFT Data Construction}
    \item \hyperref[app:use_grpo]{\ref*{app:use_grpo}. Use GRPO Reward Computation}
    \item \hyperref[app:experimental_details]{\ref*{app:experimental_details}. Additional Experimental Details}
    \item \hyperref[app:95_ci]{\ref*{app:95_ci}. Confidence Interval Analysis}
    \item \hyperref[app:more_ablations]{\ref*{app:more_ablations}. More Ablation Studies}
\end{itemize}

\vspace{0.5em}

\section{Answer Sufficiency Assessment}
\label{app:memory_sufficiency}

\subsection{Pilot Dataset Preparation}
\label{app:ap_dataset}

Starting from 580 pilot questions sampled from WorldMemArena (WMA)~\cite{liu2026worldmemarena}, we first evaluate each question without providing memory and exclude 45 questions answered correctly under this setting, leaving 535 questions for further assessment.
We consider four memory conditions: Gold Evidence, AUGUSTUS~\cite{jain2025augustus}, Omni-SimpleMem~\cite{liu2026omni}, and UniversalRAG~\cite{yeo2026universalrag}.
Gold Evidence consists of human-annotated oracle evidence provided by WorldMemArena, while the other three memory conditions are constructed using Qwen3-VL-4B-Instruct.

\subsection{LLM-Based Sufficiency Assessment}
\label{app:ap_judge}

For each memory condition, we replace the original images with dataset-provided reference captions while preserving the textual memory content.
This allows the judge to assess the availability of relevant visual and textual information through a text-based representation of the memory.
We employ GPT-6-Luna~\cite{openai2026gpt6luna} as an LLM judge to determine whether each memory contains sufficient information to answer the corresponding question.
Given the original question, gold answer, and reference-caption memory, the judge assigns one of three labels: \textit{full}, \textit{partial}, or \textit{insufficient}.
A memory is labeled \textit{full} if all information necessary to determine the gold answer is available, \textit{partial} if only some required information is present, and \textit{insufficient} if the necessary information is absent or only tangentially related.
The judge assesses information availability rather than predicting whether a VLM can correctly answer the question.
The complete judge prompt is provided in Table~\ref{box:ap_prompt}.

\begin{table}[t]
\centering
\begin{tcolorbox}[
    boxrule=0.5pt,
    arc=1mm,
    left=1.5mm,
    right=1.5mm,
    top=1mm,
    bottom=1mm,
    title={Answer Sufficiency Judge Prompt},
    fonttitle=\bfseries\footnotesize,
]
{
\scriptsize
\raggedright

You are auditing a memory system. A model was shown the MEMORY below (and nothing else from the past) and asked the QUESTION.

\smallskip
Decide whether the MEMORY contains sufficient information to determine the GOLD ANSWER. Images were replaced by their captions.

\smallskip
Grade:

\textbf{"full":} every fact needed for the gold answer is present (explicitly, or trivially derivable such as a count of listed items or a date from an explicit date + "yesterday").

\textbf{"partial":} some but not all of the needed facts are present (e.g. 3 of 5 items, the entity but not the attribute, the event but not when).

\textbf{"insufficient":} the needed facts are absent or only tangentially related.

\smallskip
Judge information content only, not whether a model would succeed.

\smallskip
QUESTION: \{question\}

GOLD ANSWER: \{answer\}

\smallskip
MEMORY:

\{memory\}

\smallskip
Return JSON only:

\{"ap": "full|partial|insufficient", "missing": "$<$what is missing, one short phrase, or none$>$"\}
}
\end{tcolorbox}
\caption{Answer Sufficiency Judge Prompt}
\label{box:ap_prompt}
\end{table}

\subsection{Answer-Sufficient Instance Selection}
\label{app:ap_selection}
Table~\ref{tab:answer_sufficiency} summarizes the assessment results across the four memory conditions.
We retain only instances labeled \textit{full} for subsequent memory utilization analyses, excluding partially sufficient, insufficient, and unsuccessful judge evaluations.
Importantly, the sufficiency assessment is performed using reference captions, whereas downstream question answering is evaluated using the original multimodal memory containing pixel images.
Since each memory condition is filtered independently, the resulting evaluation subsets differ across memory systems.
\begin{table}[t]
\centering
\footnotesize
\resizebox{\columnwidth}{!}{
\begin{tabular}{lcccc}
    \toprule
    \textbf{Memory Condition} &
    \textbf{Full} &
    \textbf{Partial} &
    \textbf{Insufficient} &
    \textbf{Failed} \\
    \midrule
    \textit{Gold Evidence}  & 349 & 181 & 4  & 1 \\
    \textit{AUGUSTUS}       & 245 & 269 & 21 & 0 \\
    \textit{Omni-SimpleMem} & 262 & 252 & 21 & 0 \\
    \textit{UniversalRAG}   & 296 & 227 & 9  & 3 \\
    \bottomrule
\end{tabular}
}
\caption{
Answer sufficiency assessment across four memory conditions.
Only instances labeled \textbf{Full} are retained for subsequent memory utilization analyses.
}
\label{tab:answer_sufficiency}
\end{table}

\section{Caption-Based Memory Construction}
\label{app:captioning}

\subsection{Caption Generation}
We conduct captioning experiments using AUGUSTUS~\cite{jain2025augustus} memories to evaluate whether model-generated textual descriptions preserve answer-relevant visual information.
We use Qwen3-VL-4B-Instruct and GPT-4o-mini as caption generators to examine whether a more capable external VLM can better preserve answer-relevant visual information.
Captions are generated without access to downstream questions.
The same caption generation prompt is used for both models, as shown in Table~\ref{box:caption_prompt}.

\begin{table}[t]
\centering
\begin{tcolorbox}[
    boxrule=0.5pt,
    arc=1mm,
    left=1.5mm,
    right=1.5mm,
    top=1mm,
    bottom=1mm,
    title={Caption Generation Prompt},
    fonttitle=\bfseries\footnotesize,
]
{\footnotesize
\raggedright
Describe this image in detail so that someone who cannot see it could answer questions about it. Mention the application or page, the main elements, any visible text, numbers, names, and the state of the interface or scene. Write 3-6 sentences.
}
\end{tcolorbox}
\caption{Caption Generation Prompt}
\label{box:caption_prompt}
\end{table}

\subsection{Caption-Based Memory Construction}
We consider two strategies for incorporating model-generated captions into multimodal memory.

\begin{itemize}
    \item \textbf{Caption Replacement :}
    Each original image is replaced with its generated caption, while preserving the image identifier.
    Specifically, \texttt{image <id>: <pixel>} is replaced with \texttt{image <id> caption: <text>}.

    \item \textbf{Caption Augmentation :}
    The generated caption is inserted immediately before the corresponding image, retaining the original pixel content.
\end{itemize}

\noindent Both strategies preserve the retrieved memory entries and their ordering, altering only the visual evidence.

\subsection{Answer Sufficiency After Caption Replacement}
To examine whether caption replacement preserves the information required for downstream question answering, we repeat the answer sufficiency assessment described in Appendix~\ref{app:memory_sufficiency}.
We evaluate 245 originally answer-sufficient AUGUSTUS memory instances after replacing their images with model-generated captions.
As shown in Table~\ref{tab:caption_sufficiency}, 60.8\% of GPT-4o-mini caption memories and 67.3\% of Qwen3-VL-4B caption memories remain answer-sufficient.
The remaining instances are classified as partially sufficient or insufficient, indicating that generated captions can omit information needed to answer the original questions.

\subsection{Qualitative Examples}
Table~\ref{tab:caption_qualitative} presents qualitative examples of answer-relevant information lost during caption replacement across GUI, embodied, and lifelong scenarios.
In the GUI example, a small loading spinner is omitted or incorrectly identified as a gear icon.
In the embodied example, both generated captions fail to identify the seeds held by the player.
In the lifelong example, the captions omit the crossed-out state of dates on a whiteboard, which is essential for answering the question.
These examples illustrate how generated captions may preserve broader scene descriptions while losing specific visual details required for downstream QA.

\begin{table}[t]
\centering
\setlength{\tabcolsep}{5pt}
\renewcommand{\arraystretch}{1.12}
\resizebox{\columnwidth}{!}{
\begin{tabular}{lcccc}
\toprule
\textbf{Caption Generator} &
\textbf{Full} &
\textbf{Partial} &
\textbf{Insufficient} &
\textbf{Failed} \\
\midrule
GPT-4o-mini &
149 (\textbf{60.8\%}) &
78 (31.8\%) &
14 (5.7\%) &
4 (1.6\%) \\
Qwen3-VL-4B &
165 (\textbf{67.3\%}) &
67 (27.3\%) &
4 (1.6\%) &
9 (3.7\%) \\
\bottomrule
\end{tabular}
}
\caption{
Answer sufficiency statistics after caption replacement on originally answer-sufficient AUGUSTUS memory instances.
}
\label{tab:caption_sufficiency}
\end{table}

\begin{table*}[t]
\centering
\setlength{\tabcolsep}{4pt}
\renewcommand{\arraystretch}{1.1}
\fontsize{6}{6.6}\selectfont
\begin{tabular}{@{}p{0.205\textwidth}p{0.765\textwidth}@{}}
\toprule
\textbf{Example} & \textbf{Caption text} \\
\midrule
\begin{minipage}[t]{\linewidth}\vspace{0pt}\raggedright\includegraphics[width=0.86\linewidth]{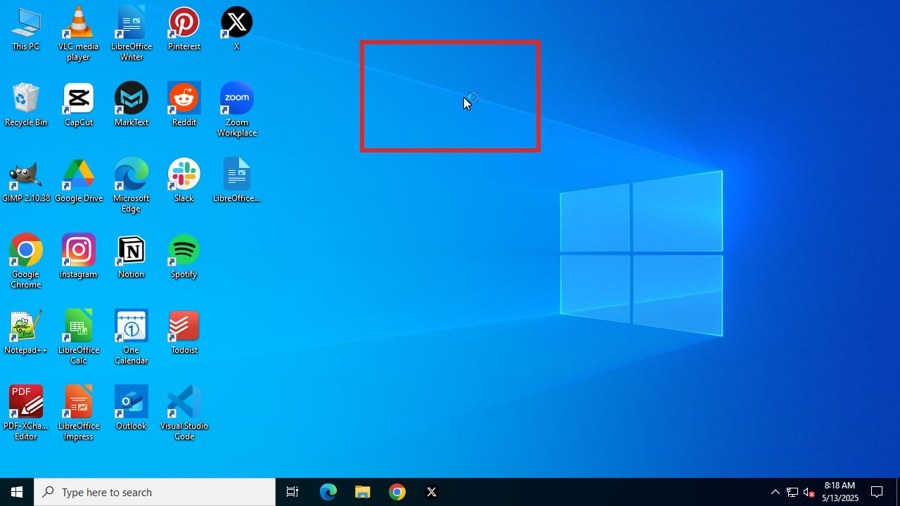}
\par\includegraphics[width=0.80\linewidth]{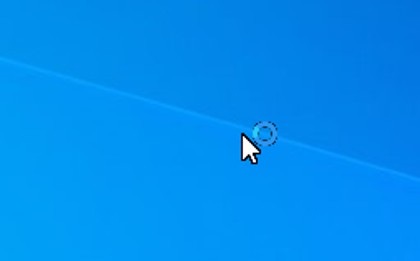}
\par\textbf{Question:} In the screenshot at step 19, what appears alongside the mouse cursor near the upper-center of the screen?
\par\textbf{Gold:} A small loading spinner appears next to the mouse cursor.\end{minipage} & 
\begin{minipage}[t]{\linewidth}
\vspace{0pt}
\setlength{\leftskip}{6pt}
\setlength{\rightskip}{6pt}
\setlength{\parskip}{3pt}
\textbf{Reference Caption}\par This is a Windows 10 desktop with the default blue wallpaper and several application shortcuts arranged in columns along the left, including Chrome, Edge, Slack, Spotify, Outlook, Notion, Zoom Workplace, CapCut, VLC, and LibreOffice apps. The taskbar is visible at the bottom with the Start/search bar and pinned apps, while \textcolor{green!45!black}{the mouse cursor with a small loading spinner appears near the upper-center of the screen}.
\par\smallskip
\hrule
\smallskip
\textbf{GPT-4o-mini Caption}\par The image displays a Windows 10 desktop environment. The background is a solid blue gradient, and the taskbar is located at the bottom of the screen, showing the time as 8:18 AM and the date as May 13, 2025. There are several application icons arranged on the desktop, including common programs like "Google Chrome," "Spotify," and "Zoom Workplace." The icons are visually distinct, featuring logos for each application, such as a green and white "Notion" logo and a colorful "GIMP" icon. The Recycle Bin is located in the top left corner, and various productivity and media applications are scattered throughout the desktop. \textcolor{red!75!black}{The cursor is positioned near the top left}, indicating potential interaction with the desktop elements.\par{\color{red!75!black}\footnotesize\emph{[The loading spinner is omitted.]}}\par\smallskip
\hrule
\smallskip
\textbf{Qwen3-VL-4B Caption}\par This image shows a Windows 10 desktop with a blue default wallpaper featuring the iconic Windows logo in the background. The desktop is populated with numerous application icons, including common tools like Google Chrome, Microsoft Edge, VLC media player, and LibreOffice suite applications, as well as social media and productivity apps such as Pinterest, Reddit, Slack, and Todoist. The taskbar at the bottom displays the search bar, system tray with icons for volume, notifications, and date/time (showing 8:18 AM, 5/13/2025), and the Start button. The cursor is positioned near the center of the screen, hovering over a \textcolor{red!75!black}{small gear icon}, likely indicating settings or system options. The desktop is organized in a grid layout with no open windows, suggesting a clean and idle state.\par{\color{red!75!black}\footnotesize\emph{[The loading spinner is misidentified as a gear icon.]}}
\vspace{3pt}
\end{minipage} \\
\midrule
\begin{minipage}[t]{\linewidth}\vspace{0pt}\raggedright\includegraphics[width=0.86\linewidth]{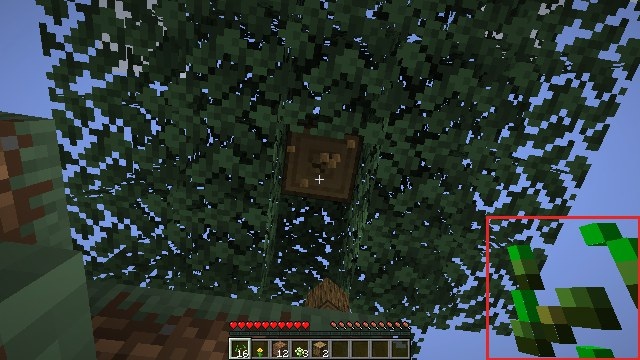}
\par\includegraphics[width=0.80\linewidth]{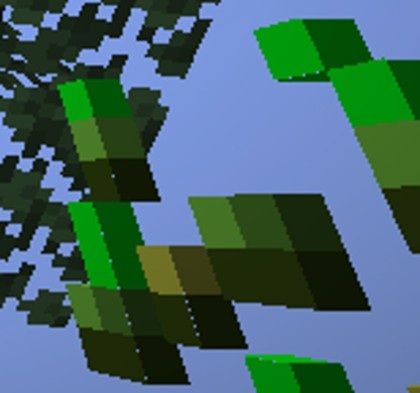}
\par\textbf{Question:} In the frame looking straight up through a dense square opening in a leafy tree canopy with blue sky in the center, what item is the player holding?
\par\textbf{Gold:} The player is holding green seeds.\end{minipage} & 
\begin{minipage}[t]{\linewidth}
\vspace{0pt}
\setlength{\leftskip}{6pt}
\setlength{\rightskip}{6pt}
\setlength{\parskip}{3pt}
\textbf{Reference Caption}\par A first-person Minecraft view looks up into the leafy canopy of a tree, centering on a wood log block overhead with another trunk segment visible lower down. The hotbar and hearts/hunger UI are visible at the bottom, and \textcolor{green!45!black}{the player appears to be holding seeds in their hand on the right side of the screen}.\par\smallskip
\hrule
\smallskip
\textbf{GPT-4o-mini Caption}\par The image is taken from the game Minecraft, showing a view from below a tree. The player is looking up at the underside of a tree canopy, where a wooden log block is visible among the dense green leaves. The scene is set against a clear blue sky. In the bottom portion of the screen, there is a user interface displaying the player's inventory, which includes 10 pieces of an item (\textcolor{red!75!black}{likely wood}), 12 pieces of another item (\textcolor{red!75!black}{possibly leaves}), and 2 of a third item. The health and hunger bars are also visible, indicating the player's current status in the game. The overall perspective gives a sense of being directly beneath the tree, emphasizing the height and density of the foliage above.\par{\color{red!75!black}\footnotesize\emph{[The item held in the player's hand (green seeds) is omitted.]}}\par\smallskip
\hrule
\smallskip
\textbf{Qwen3-VL-4B Caption}\par This image is a screenshot from the popular sandbox game Minecraft, showing a first-person view of a player looking upward at a large, dense tree with green leaves and a wooden sign hanging from its branches. The player's health and hunger bars are visible at the bottom of the screen, along with a hotbar containing various items, including \textcolor{red!75!black}{16 sticks, 12 wood, and 2 green blocks}. The sky is a clear blue, and the scene appears to be set in a forest biome. The sign is centered in the tree canopy, but no readable text or names are visible on it. The player's inventory and UI elements suggest they are actively exploring or gathering resources.\par{\color{red!75!black}\footnotesize\emph{[The held seeds are omitted; inventory items are misidentified.]}}\vspace{3pt}\end{minipage} \\
\midrule
\begin{minipage}[t]{\linewidth}\vspace{0pt}\raggedright\includegraphics[width=0.86\linewidth]{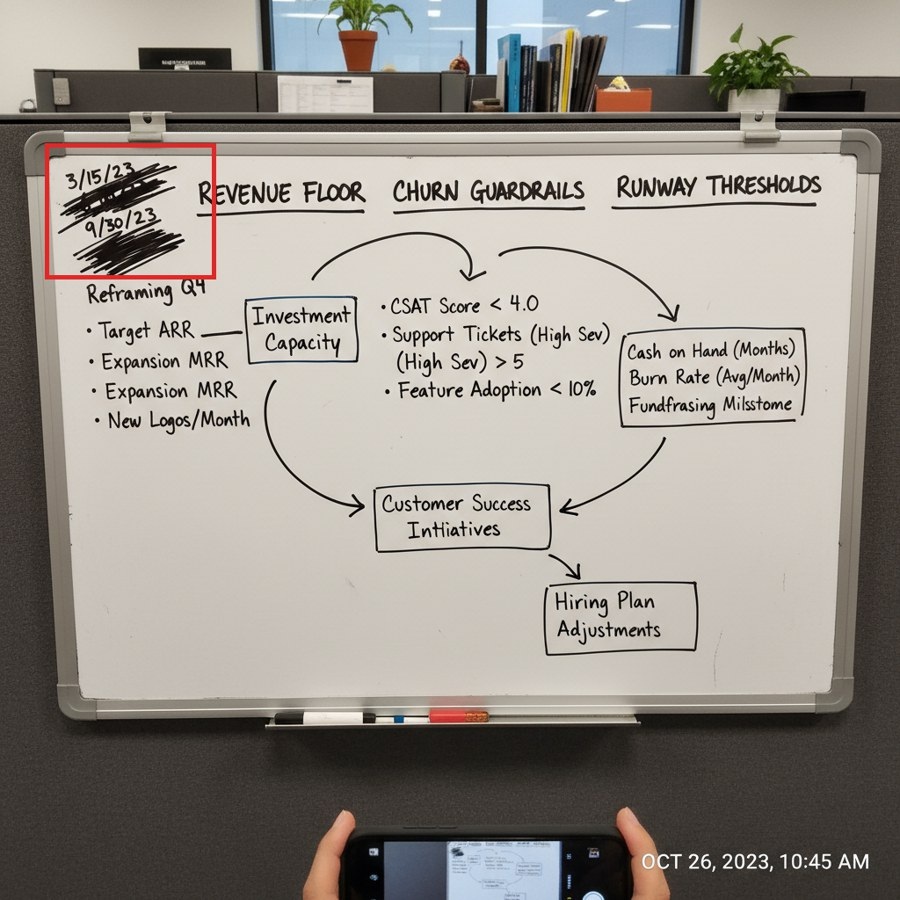}
\par\includegraphics[width=0.80\linewidth]{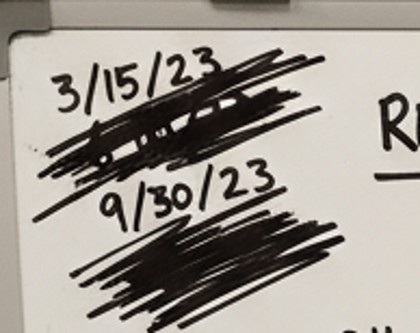}
\par\textbf{Question:} In the quit-decision whiteboard image, what visual clue shows the framework was still being worked through, not presented as a settled plan?
\par\textbf{Gold:} The crossed-out resignation dates in the corner show that the decision was still being revised.\end{minipage} & \begin{minipage}[t]{\linewidth}
\vspace{0pt}
\setlength{\leftskip}{6pt}
\setlength{\rightskip}{6pt}
\setlength{\parskip}{3pt}
\textbf{Reference Caption}\par A phone photo of Maya's whiteboard with three columns labeled revenue floor, churn guardrails, and runway thresholds, plus arrows showing which metrics feed each decision. \textcolor{green!45!black}{There are a few crossed-out resignation dates in the corner}, which makes it clear this was a real reframing moment rather than a neat final framework.\par\smallskip
\hrule
\smallskip
\textbf{GPT-4o-mini Caption}\par The image shows a whiteboard filled with handwritten notes and diagrams, likely from a business meeting or brainstorming session. The main sections are labeled "Revenue Floor," "Churn Guardrails," and "Runway Thresholds." Under "Revenue Floor," there are bullet points including "Target ARR," "Expansion MRR," and "New Logos/Month." The "Churn Guardrails" section includes criteria such as "CSAT Score \textless{} 4.0," "Support Tickets (High Sev) \textgreater{} 5," and "Feature Adoption \textless{} 10\%." The "Runway Thresholds" section lists "Cash on Hand (Months)," "Burn Rate (Avg/Month)," and "Fundraising Milestone." The date and time at the bottom indicate it was captured on October 26, 2023, at 10:45 AM, and a hand holding a smartphone is visible at the bottom of the image, suggesting the photo was taken to document the whiteboard content.\par{\color{red!75!black}\footnotesize\emph{[The crossed-out dates in the corner are omitted.]}}\par\smallskip
\hrule
\smallskip
\textbf{Qwen3-VL-4B Caption}\par The image shows a whiteboard in an office setting, with a hand holding a smartphone at the bottom, capturing a photo of the board dated October 26, 2023, at 10:45 AM. The whiteboard is divided into three main sections: "REVENUE FLOOR," "CHURN GUARDRAILS," and "RUNWAY THRESHOLDS," each with associated metrics and arrows connecting them to "Investment Capacity," "Customer Success Initiatives," and "Hiring Plan Adjustments." The top left corner has \textcolor{red!75!black}{scribbled dates} (3/15/23, 8/10/23, 9/30/23) and the note "Reframing Q4" with bullet points about ARR, MRR, and logos. The board's content is handwritten in black marker, with some text underlined, and the phone's camera interface is visible at the bottom of the frame.\par{\color{red!75!black}\footnotesize\emph{[Dates are mentioned, but the crossed-out state is omitted.]}}\end{minipage} \\
\bottomrule
\end{tabular}
\caption{Qualitative examples of information loss during caption replacement across GUI, embodied, and lifelong scenarios. \textcolor{green!45!black}{Green text} highlights answer-relevant evidence retained in reference captions; \textcolor{red!75!black}{red text} highlights incorrect generated descriptions or editorial notes identifying omissions. All captions are reproduced in full text.}
\label{tab:caption_qualitative}
\end{table*}

\section{Read Probe Construction}
\label{app:read_use_diagnosis}

\subsection{Answer-Critical Fact Extraction and Probe Generation}
\label{app:fact_extraction}
We construct read probes to assess whether VLMs can accurately recover information needed to answer downstream questions.
Probe construction is performed only on instances classified as fully answer-sufficient in Appendix~\ref{app:memory_sufficiency}.
For each instance, GPT-6-Luna receives the original question, gold answer, and the same reference-caption memory used for answer sufficiency assessment.
We prompt the model to identify up to four answer-critical facts necessary to determine the gold answer.
For each fact, the model generates a source label (\textit{text} or \textit{image}), an evidence locator, a probe question, and a short reference answer of one to eight words.
Each probe is designed to assess a single fact without requiring counting, comparison, or reasoning across multiple facts.
This allows us to assess evidence reading separately from downstream reasoning.
We further validate whether the extracted facts collectively suffice to answer the original questions in Appendix~\ref{app:fact_sufficiency_validation}.
The complete fact extraction and probe generation prompt is provided in Table~\ref{box:fact_extraction}.
\begin{table}[t]
\centering

\begin{tcolorbox}[
    boxrule=0.5pt,
    arc=1mm,
    left=1.5mm,
    right=1.5mm,
    title={Answer-Critical Fact Extraction Prompt},
    fonttitle=\bfseries\footnotesize,
]
{
\scriptsize
\raggedright
You are building reading-comprehension probes for a memory audit. A model was shown the MEMORY below and asked the QUESTION.

\smallskip
The MEMORY has been verified to contain enough information to determine the GOLD ANSWER. Images were replaced by their captions (written as [image <id>: <caption>]); the model itself saw the actual image.

\smallskip
Task: list the MINIMAL set of answer-critical facts --- the facts in the MEMORY without which the gold answer cannot be determined (at most 4; merge facts that come from the same sentence). Cover every component the gold answer depends on, including dates or times that appear only in memory headers or timestamps.

\smallskip
For each fact write ONE probe question that:

\begin{itemize}
    \setlength{\itemsep}{0pt}
    \setlength{\topsep}{0pt}
    \item checks only whether the model READ that single fact from the memory;
    \item includes a locator cue taken from the memory;
    \item has a short answer (1--8 words) stated in or directly visible in the memory;
    \item does not require counting, comparing, or combining several facts.
\end{itemize}

Do not invent facts that are not in the MEMORY.

\smallskip
QUESTION: \{question\}

GOLD ANSWER: \{answer\}

MEMORY:

\{memory\}

\smallskip
Return JSON only:

\{"facts": [\{"fact": "...", "source": "text|image", "locator": "...", "probe": "...", "probe\_answer": "..."\}]\}
}
\end{tcolorbox}
\caption{Answer-Critical Fact Extraction Prompt}
\label{box:fact_extraction}
\end{table}

\subsection{Locator-Free Probe Refinement}
\label{app:locator_removal}
Initially generated probes contain explicit location cues, such as memory indices, session identifiers, and image IDs, which directly indicate where the relevant evidence is stored.
To avoid providing these cues during evaluation, we rewrite each probe using GPT-6-Luna to identify the target evidence through its content or context instead.
The rewriting process removes explicit location references while preserving the original fact and expected answer.
We further check the rewritten probes for remaining location cues using regular-expression matching and repeat the rewriting process up to three times when such cues are detected.
All subsequent read evaluations use these locator-free probes.
The rewriting prompt is provided in Table~\ref{box:locator_removal}, and Table~\ref{tab:locator_examples} presents representative examples.
\begin{table}[t]
\centering
\begin{tcolorbox}[
    boxrule=0.5pt,
    arc=1mm,
    left=1.5mm,
    right=1.5mm,
    title={Locator-Free Probe Rewriting Prompt},
    fonttitle=\bfseries\footnotesize,
]
{\scriptsize\raggedright
You are rewriting reading-comprehension probes so that they do NOT tell the reader where to look.

\smallskip
Each probe below checks one fact from the MEMORY, but it currently contains a location cue (e.g. "In Memory 5", "turn D2:6", "session D3", "image D5:IMG\_002", "the header", "step 9", "the last note").

\smallskip
Rewrite every probe so that:

\begin{itemize}
    \setlength{\itemsep}{0pt}
    \setlength{\topsep}{0pt}
    \item it contains NO location cue of any kind;
    \item it identifies the target by CONTENT or context instead, precisely enough that only one answer fits given the MEMORY;
    \item it never contains or hints at its own answer, and still asks for the same fact with the same short answer;
    \item dates are identified by the content of the conversation or event, rather than a memory header.
\end{itemize}

\smallskip
MEMORY (for disambiguation only):

\{memory\}

\smallskip
PROBES (JSON): \{probes\}

\smallskip
Return JSON only:

\{"probes": ["<rewritten probe 1>", "<rewritten probe 2>", ...]\} in the same order.
}
\end{tcolorbox}
\caption{Prompt for Locator-Free Probe Rewriting}
\label{box:locator_removal}
\end{table}

\begin{table*}[t]
\centering
\footnotesize
\setlength{\tabcolsep}{5pt}
\renewcommand{\arraystretch}{1.15}

\begin{tabular}{
    p{0.3\textwidth}
    p{0.3\textwidth}
    p{0.3\textwidth}
}
\toprule
\textbf{Instance ID} &
\textbf{Locator-Cued Probe} &
\textbf{Locator-Free Probe} \\
\midrule


\multirow{4}{*}{
\shortstack[l]{
\texttt{web\_01::f017f95f::q10}
}}
&
What message did
\textcolor{red!75!black}{image web\_01\_img\_002}
show on reddit.com?
&
What message appeared on
\textcolor{green!45!black}{the Reddit page the assistant tried to access}?
\\

\cmidrule(lr){2-3}

&
What URL did the agent type in
\textcolor{red!75!black}{Memory 3}?
&
What address did the assistant enter
\textcolor{green!45!black}{to open a meme subreddit on Reddit}?
\\

\midrule


\multirow{6}{*}{
\shortstack[l]{
\texttt{eb\_alfred\_complex\_instruction\_02}\\
\texttt{::c9b5dbd0::q09}
}}
&
In
\textcolor{red!75!black}{Memory 3's image},
what gray object is visible in the foreground?
&
In
\textcolor{green!45!black}{the scene with a pink-stoned ring on a wooden shelf and a dark blue book at the right},
what gray object is visible in the foreground?
\\

\cmidrule(lr){2-3}

&
In
\textcolor{red!75!black}{Memory 1},
what action did the robot take with the object in hand?
&
\textcolor{green!45!black}{After the robot couldn't pick up the watch because it was already holding the bowl},
what action did it take?
\\

\bottomrule
\end{tabular}

\caption{
\textbf{Examples of locator-free probe refinement.}
Explicit location cues (\textcolor{red!75!black}{red})
are replaced with content-based references
(\textcolor{green!45!black}{green}),
while preserving the queried facts and expected answers.
Examples are drawn from GUI and embodied agent tasks.
}
\label{tab:locator_examples}
\end{table*}

\subsection{Read and Use Evaluation}
\label{app:read_use_evaluation}
To assess evidence reading and downstream utilization, we evaluate VLMs using the same multimodal memory containing pixel images.
For evidence reading, the model answers each locator-free read probe independently, whereas downstream utilization is evaluated using the original question.
Both evaluations follow the same memory-based question-answering setup, with different instructions for short factual answers and complete downstream answers, illustrated in Table~\ref{box:read_use_eval}.
We use greedy decoding with a maximum of 512 generated tokens and disable thinking during inference.
Model responses are evaluated by GPT-6-Luna, which assigns scores of 0, 0.25, 0.5, 0.75, or 1.
For text-based read probes, a response passes if its discretized judge score equals 1.
For image-based read probes, we accept a response if its raw judge score is at least 0.75 or its token-level F1 score is at least 0.5, allowing for lexical differences between reference-caption-derived answers and responses grounded in the original images.

An instance is considered successfully read only if all its generated read probes pass.
Formally, given a question $q$ and memory $M$ with $K_q$ read probes, we define
\begin{equation}
    \mathrm{Read}(q,M)
    =
    \prod_{i=1}^{K_q}
    \mathbb{I}\left[\mathrm{Pass}(p_i,M)\right],
\end{equation}
where $\mathrm{Pass}(p_i,M)$ indicates whether the model correctly answers probe $p_i$ under its source-specific evaluation criterion.
For downstream utilization, we define $\mathrm{Use}(q,M)=1$ when the judge score for the original question equals 1.
\begin{table}[t]
\centering

\begin{tcolorbox}[
    boxrule=0.5pt,
    arc=1mm,
    left=1.5mm,
    right=1.5mm,
    title={Read/Use Evaluation Prompt},
    fonttitle=\bfseries\footnotesize,
]
{\scriptsize\raggedright
You are given relevant memory from earlier interactions and a current question.

\smallskip
The memory has already been retrieved for you.

Use the memory when it is relevant to solving the current question.

Do not assume information that is not supported by the supplied memory or question.

\smallskip
Relevant memory:

\{memory\}

\smallskip
\textbf{[Read evaluation]}

Question:

\{probe\}

Answer concisely with only the requested information.

\smallskip
\textbf{[Use evaluation]}

Current question:

\{question\}

\{question\_image\}

Answer completely: include every step, entity, or detail needed to fully answer the question.
}
\end{tcolorbox}

\caption{Read/Use Evaluation Prompt}
\label{box:read_use_eval}
\end{table}
\begin{table*}[t]
\centering

\begin{tcolorbox}[
    boxrule=0.5pt,
    arc=1mm,
    left=1.5mm,
    right=1.5mm,
    title={Use GRPO Correctness Reward Judge Prompt},
    fonttitle=\bfseries,
]
{\footnotesize
\raggedright

You are an impartial judge evaluating the memory capabilities of an AI assistant with the question-answering task.
Your task is to compare the Assistant's Answer against the Ground Truth and assign a score of 0, 0.25, 0.5, 0.75, or 1.

\medskip
\textbf{Scoring Rubric}

\medskip
\textbf{Score 0 (Incorrect / Miss):}

- The answer contradicts the Ground Truth.

- For Yes/No questions: The answer has the wrong polarity (e.g., says ``Yes'' when Ground Truth is ``No'').

- For Open-ended questions: The answer provides factually wrong information or hallucinations.

- The assistant fails to provide the required information.

\medskip
\textbf{Score 0.25 (Poor / Tangential):}

- The answer touches on the topic but misses the \textbf{core entity} or key value required.

- The answer contains a mix of minor correct details and \textbf{significant hallucinations} or wrong associations.

- The answer is excessively vague to the point of being useless (e.g., answering ``a dog'' instead of ``a golden retriever'').

\medskip
\textbf{Score 0.5 (Partial / Vague):}

- The answer is technically correct, but lacks confidence or is incomplete.

- The answer captures the \textbf{main entity or concept} correctly but misses a part of the required supporting details.

- For Yes/No questions: The polarity is correct, but the reasoning is flawed (if have), or the assistant is uncertain (e.g., ``I think it might be Yes'').

- For Open-ended questions: The answer is too general or misses key adjectives/details present in the Ground Truth.

\medskip
\textbf{Score 0.75 (Good / Minor Imperfection):}

- The answer is largely accurate and captures the core information confidently.

- It misses only \textbf{minor details} (e.g., specific adjectives or secondary details) that do not alter the main truth.

- The answer contains all the correct information but includes unnecessary ``fluff'' or slight conversational filler that reduces precision.

\medskip
\textbf{Score 1 (Correct / Exact):}

- The answer is accurate, precise, and confident.

- For Yes/No questions: The polarity matches the Ground Truth perfectly.

- For Open-ended questions: The answer contains \textbf{all} the core information and necessary details required by the Ground Truth without hallucinations.

\medskip
\textbf{Input Data}

Question: \texttt{\{\{question\}\}}

Ground Truth: \texttt{\{\{ground\_truth\}\}}

Assistant Answer: \texttt{\{\{model\_output\}\}}

\medskip
\textbf{Output Format}

Output strictly in the following JSON format:

\{"score": \textless 0, 0.25, 0.5, 0.75, or 1\textgreater, "reasoning": "\textless short explanation\textgreater"\}}

\end{tcolorbox}

\caption{Use GRPO Correctness Reward Judge Prompt}
\label{box:use_reward_judge}
\end{table*}
\begin{table}[t]
\centering
\small
\resizebox{\columnwidth}{!}{
\begin{tabular}{ll}
    \toprule
    \textbf{Evaluation} & \textbf{Success Criterion} \\
    \midrule
    Text Read
    & Discretized judge score $=1$ \\
    Image Read
    & Raw judge score $\geq 0.75$ or token-F1 $\geq 0.5$ \\
    Overall Read
    & All generated read probes pass \\
    Use
    & Discretized judge score $=1$ \\
    \bottomrule
\end{tabular}
}
\caption{
Evaluation criteria for read probes and downstream question answering.
}
\label{tab:read_use_criteria}
\end{table}

\subsection{Failure Categorization}
\label{app:failure_categorization}
Based on the outcomes of $\mathrm{Read}(q,M)$ and $\mathrm{Use}(q,M)$, we categorize each instance into four groups:
\begin{itemize}
    \item \textbf{Read Failure}: Both evidence reading and downstream QA fail ($\mathrm{Read}=0$, $\mathrm{Use}=0$).
    \vspace{1mm}
    \item \textbf{Use Failure}: Evidence reading succeeds, but downstream QA fails ($\mathrm{Read}=1$, $\mathrm{Use}=0$).
    \vspace{1mm}
    \item \textbf{QA Success (Read \ding{55})}: Downstream QA succeeds despite failed evidence reading ($\mathrm{Read}=0$, 
    $\mathrm{Use}=1$).
    \vspace{1mm}
    \item \textbf{QA Success (Read \ding{51})}: Both evidence reading and downstream QA succeed ($\mathrm{Read}=1$, $\mathrm{Use}=1$).
\end{itemize}

\subsection{Human Validation}
\label{app:human_validation}
To validate our Read/Use diagnosis, one author manually evaluated 120 stratified LLM-judge decisions and 76 generated read probes from WMA.
We annotated judge decisions as Correct, Partial, or Incorrect without revealing the LLM verdicts.
For agreement analysis, human judgments were considered successful only when labeled Correct for Use and text Read, or Correct or Partial for image Read, reflecting the more permissive image evaluation criteria.
Read probes were assessed for answer-criticality, reference correctness, unambiguity, and answer leakage.
We report population-weighted rates with 95\% confidence intervals estimated using 5,000 stratified bootstrap replicates.

As shown in Table~\ref{tab:human_validation}, the LLM judge achieves 95.1\% agreement with human judgments ($\kappa=0.83$).
Of the 10 instances where the LLM judge disagrees with human judgments, seven reflect stricter judge decisions, suggesting that the judge does not systematically favor accepting incorrect answers.
Read probes exhibit high reference correctness (92.5\%) and unambiguity (94.5\%).
The lower criticality rate (84.9\%), particularly for text-based probes, indicates that probe errors arise more often from extracting unnecessary facts than from incorrect references.
Under our all-probes-correct criterion, such over-extraction may underestimate reading success.
Importantly, the Read/Use distinction is more informative for questions requiring multiple facts.
For longer-answer questions, each instance contains 2.8 read probes on average, and 37\% of instances exhibit successful reading but unsuccessful downstream answering.
This suggests that the observed Read/Use gap is not merely an artifact of asking equivalent questions.

\begin{table}[t]
\centering
\small
\setlength{\tabcolsep}{5pt}
\renewcommand{\arraystretch}{1.12}
\begin{tabular}{lcc}
\toprule
\textbf{Validation} & \textbf{Rate (\%)} & \textbf{95\% CI} \\
\midrule
\rowcolor{gray!10}\multicolumn{3}{l}{\textit{LLM Judge (120 decisions)}} \\
Human agreement & 95.1 & [90.4, 98.4] \\
\midrule
\rowcolor{gray!10}\multicolumn{3}{l}{\textit{Read Probes (76 probes)}} \\
Criticality & 84.9 & [74.2, 94.5] \\
Reference correctness & 92.5 & [84.9, 100.0] \\
Unambiguity & 94.5 & [87.7, 99.5] \\
No answer leakage & 95.4 & [88.6, 100.0] \\
\bottomrule
\end{tabular}
\caption{
\textbf{Human validation of the LLM judge and read probes.}
Population-weighted rates (\%) with 95\% stratified bootstrap confidence intervals.
}
\label{tab:human_validation}
\end{table}

\subsection{Answer-Critical Fact Sufficiency Validation}
\label{app:fact_sufficiency_validation}
To validate whether the extracted answer-critical facts collectively preserve sufficient information for downstream question answering, we conduct a fact-only QA evaluation.
Specifically, we replace the original multimodal memory with the extracted answer-critical facts and ask Qwen3-VL-4B-Instruct, the same model used in our Read/Use diagnosis, to answer the original questions.
The model achieves 97.6\% accuracy in this setting, suggesting that the extracted facts preserve sufficient information to answer the vast majority of evaluated questions.
We further examine Use Failures instances, where the model correctly answers all read probes but fails the original downstream question.
When the original memory is replaced with the extracted facts, the model correctly answers approximately 89\% of these previously failed questions.
This recovery persists for both single-fact (93.3\%) and multi-fact (84.8\%) instances, indicating that the result is not limited to questions involving a single piece of evidence.
These findings support the collective sufficiency of the extracted answer-critical facts and suggest that many observed Use Failures arise from difficulties in utilizing evidence within the original multimodal memory context, rather than an inability to solve the questions when the relevant facts are explicitly provided.

\subsection{Qualitative Read Probe Examples}
\label{app:read_examples}

Table~\ref{tab:qualitative_read_use} presents two qualitative examples from GUI and embodied agent tasks to illustrate the distinction between evidence reading and downstream utilization.
In the GUI example, the model correctly identifies the network security block from visual memory and the alternative Reddit URL from textual memory, but fails to determine the appropriate next action.
Similarly, in the embodied example, the model correctly identifies the gray bowl in the visual observation and the action of putting down the held object, yet incorrectly infers the agent's subsequent held-item state.
In both cases, the model successfully answers all evidence-specific read probes but fails the original downstream question, illustrating that successful evidence reading does not necessarily translate into effective memory utilization.

\begin{table*}[t]
\centering
\footnotesize
\setlength{\tabcolsep}{6pt}
\renewcommand{\arraystretch}{1.25}

\begin{tabular}{
    p{0.25\textwidth}
    p{0.37\textwidth}
    p{0.30\textwidth}
}
\toprule
\textbf{Visual Evidence} &
\textbf{Read Probes} &
\textbf{Downstream QA} \\
\midrule


\begin{minipage}[t]{\linewidth}
\textbf{GUI / AUGUSTUS}

\vspace{2mm}
\includegraphics[
    width=\linewidth,
    keepaspectratio
]{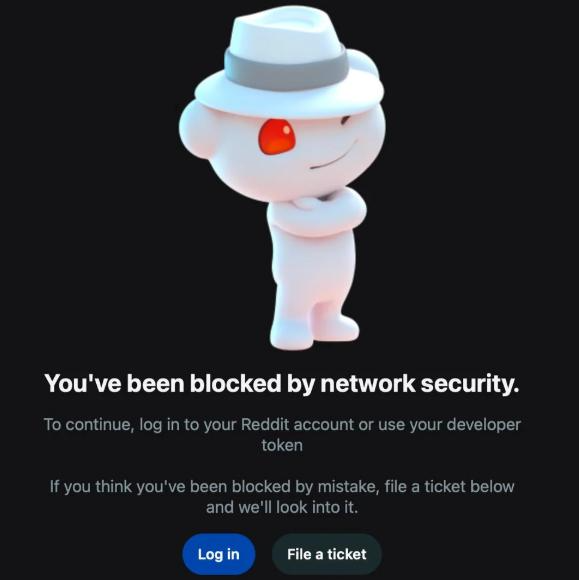}
\end{minipage}
&
\begin{minipage}[t]{\linewidth}
\textbf{Image Probe \cmark}

What message appeared on the Reddit page
the assistant tried to access?

\medskip

\textit{Expected:} Network security block

\textit{Predicted:} You've been blocked
by network security.

\par\vspace{1mm}
\noindent\hrule
\par\vspace{2mm}
\noindent\textbf{Text Probe \cmark}

What address did the assistant enter
to open a meme subreddit on Reddit?

\medskip

\textit{Expected:} old.reddit.com/r/memes

\textit{Predicted:} old.reddit.com/r/memes
\end{minipage}
&
\begin{minipage}[t]{\linewidth}
\textbf{Original Question \xmark}

What should happen next after the Reddit
security block and alternate URL entry?

\medskip

\textit{Gold:} Submit the alternate URL.

\textit{Predicted:} Click ``Log in'' or
``File a ticket.''
\end{minipage}
\\

\midrule


\begin{minipage}[t]{\linewidth}
\textbf{Embodied / Omni-SimpleMem}

\vspace{2mm}
\includegraphics[
    width=\linewidth,
    keepaspectratio
]{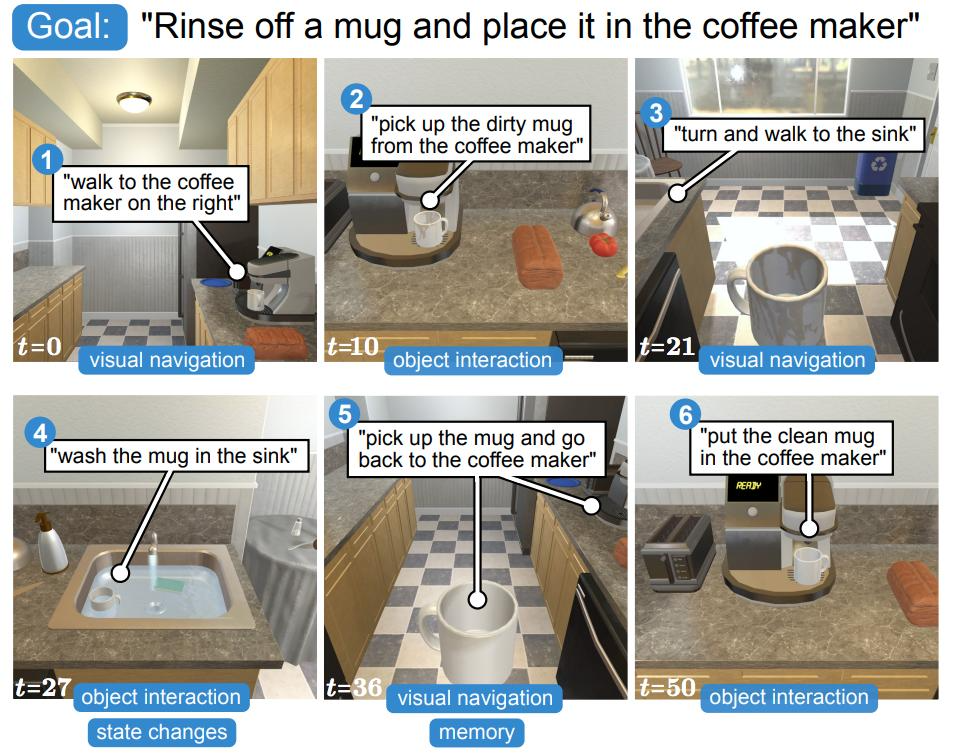}
\end{minipage}
&
\begin{minipage}[t]{\linewidth}
\textbf{Image Probe \cmark}

In the scene with a pink-stoned ring
on a wooden shelf and a dark blue book,
what gray object is visible?

\medskip

\textit{Expected:} A gray bowl

\textit{Predicted:} bowl

\par\vspace{1mm}
\noindent\hrule
\par\vspace{2mm}
\noindent\textbf{Text Probe \cmark}

After the robot could not pick up
the watch because it was holding
the bowl, what action did it take?

\medskip

\textit{Expected:} Put it down

\textit{Predicted:} put down the object in hand
\end{minipage}
&
\begin{minipage}[t]{\linewidth}
\textbf{Original Question \xmark}

What held-item state changed
between the two shelf views?

\medskip

\textit{Gold:} Holding bowl
$\rightarrow$ empty-handed.

\textit{Predicted:} Holding bowl
$\rightarrow$ holding watch.
\end{minipage}
\\

\bottomrule
\end{tabular}
\caption{
Qualitative examples of evidence-specific read probes.
Both examples successfully answer all read probes but fail downstream question answering.
}
\label{tab:qualitative_read_use}
\end{table*}

\section{Read SFT Data Construction}
\label{app:read_sft}

\subsection{Training Data Preparation}
We construct the Read SFT dataset from the training partition of our scenario-level WMA split, separately from the pilot instances used for failure diagnosis.
After filtering for supported question types, valid answers, and fully linked Gold Evidence, we generate image-based and text-based read probes following the procedure described in Appendix~\ref{app:read_use_diagnosis}, allowing multiple probes per question.

\subsection{Response Selection}
For each read probe, the original VLM generates a response using reference-caption memory, following Section~\ref{sec:read_sft}.
We retain responses that receive a correctness score of 1 from GPT-6-Luna and exclude incomplete or empty responses, as well as those that explicitly reference captions or descriptions.
Each retained response is paired with its corresponding probe and original pixel-based memory to form an SFT training example.

\subsection{Balanced Training Data}
To balance supervision across visual and textual evidence, we retain all eligible image-based examples and randomly sample an equal number of eligible text-based examples.
This yields a modality-balanced dataset of probe-level examples paired with the original multimodal memory.

\section{Use GRPO Reward Computation}
\label{app:use_grpo}

\paragraph{Correctness Reward.}
We evaluate each generated response using gpt-oss-20b~\cite{agarwal2025gpt} as an LLM judge, which shows high agreement with GPT-6-Luna used for evaluation (85.7\% agreement; Cohen's $\kappa=0.70$ on 300 responses).
Given the original question, gold answer, and generated response, the judge assigns a correctness score from $\{0, 0.25, 0.5, 0.75, 1\}$.
The score is used directly as the correctness reward $s(q,y,\hat{y})$, without binarization.
Empty responses or responses for which judging fails are assigned a score of 0.

\paragraph{Length Penalty.}
Inspired by the soft overlong penalty in DAPO~\cite{yu2026dapo}, we introduce a gradually increasing penalty to discourage excessively long responses.
Rather than using a fixed length threshold, we scale the threshold by the gold-answer length to accommodate varying answer lengths across questions.
Let $L$ denote the number of words in the generated answer and $G$ the number of words in the gold answer, with $G \geq 1$.
The penalty is defined as
\begin{equation}
\ell(\hat{y},y)
=
-\frac{1}{2}
\min\left(
1,
\frac{\max(0,L-2G)}{2G}
\right).
\end{equation}
The penalty is zero when the generated answer is no longer than twice the gold-answer length, decreases linearly beyond this threshold, and is capped at $-0.5$.
The final reward is the sum of the correctness score and length penalty, ranging from $-0.5$ to $1$.

\section{Additional Experimental Details}
\label{app:experimental_details}

\subsection{Dataset Preparation}
\label{app:dataset_preparation}

\subsubsection{Scenario-Level Split}
We use the full WorldMemArena (WMA)~\cite{liu2026worldmemarena} dataset, consisting of 461 scenarios and 24,258 question-answer pairs across GUI, embodied, and lifelong interaction domains.
To prevent scenario overlap between training and evaluation, we partition the scenarios into training, validation, and test sets with a 70:10:20 ratio using a fixed random seed of 42.
Scenarios are shuffled within their respective subfolders and assigned using a repeating 7:1:2 allocation pattern across subfolders.
This yields 323 training, 46 validation, and 92 test scenarios.

\subsubsection{Question Filtering}
We retain questions belonging to six categories: Visual Factual Recall (VFR), Visual Update (VU), Cross-Modal Reasoning (CMR), Temporal Reasoning (TR), Knowledge Reasoning (KR), and Factual Reasoning (FR).
We exclude questions with unavailable answers and require valid links to Gold Evidence, including the availability of referenced image files.
After filtering, the dataset contains 7,371 training, 1,075 validation, and 2,073 test questions.
Table~\ref{tab:wma_split_stats} summarizes the resulting distribution across domains.

\subsubsection{Stage-Specific Training Data}
We further filter the 7,371 training questions to retain only those whose Gold Evidence contains at least one image, yielding 6,722 QA instances used for Use GRPO.
For Read SFT, we generate exactly one read probe per instance, balancing image-based and text-based probes at a 1:1 ratio (3,361 each).
This results in a balanced Read SFT dataset comprising 6,722 probe-level examples.
Importantly, we use the same set of QA instances for Read SFT and the subsequent Use SFT stage.
By controlling for differences in data selection, we ensure consistent and fair comparisons across training stages and subsequent analyses.
Table~\ref{tab:training_data_stats} summarizes the training data statistics.

\begin{table}[t]
\centering
\footnotesize
\renewcommand{\arraystretch}{1.1}
\resizebox{\columnwidth}{!}{
\begin{tabular}{@{}lrrrrrr@{}}
\toprule
& \multicolumn{3}{c}{\textbf{Scenarios}}
& \multicolumn{3}{c}{\textbf{Filtered Questions}} \\
\cmidrule(lr){2-4}
\cmidrule(l){5-7}
\textbf{Domain}
& \textbf{Train} & \textbf{Val} & \textbf{Test}
& \textbf{Train} & \textbf{Val} & \textbf{Test} \\
\midrule
GUI
& 143 & 20 & 40
& 3,053 & 444 & 828 \\
Embodied
& 154 & 22 & 44
& 3,548 & 511 & 1,007 \\
Lifelong
& 26 & 4 & 8
& 770 & 120 & 238 \\
\midrule
\textbf{Total}
& \textbf{323} & \textbf{46} & \textbf{92}
& \textbf{7,371} & \textbf{1,075} & \textbf{2,073} \\
\bottomrule
\end{tabular}
}
\caption{
\textbf{WMA scenario-level split statistics.}
Numbers of scenarios and filtered questions across the training, validation, and test partitions.
}
\label{tab:wma_split_stats}
\end{table}

\begin{table}[t]
\centering
\renewcommand{\arraystretch}{1.12}
\small
\begin{tabular}{@{}lr@{}}
\toprule
\textbf{Dataset / Stage} & \textbf{Count} \\
\midrule
\multicolumn{2}{@{}l}{\emph{Data Preparation}} \\
Filtered WMA Train & 7,371 \\
Image-Containing QA & 6,722 \\
\midrule
\multicolumn{2}{@{}l}{\emph{Training}} \\
Read SFT Probes (Image / Text)
& \textbf{3,361 / 3,361} \\
Use GRPO Instances & 6,722 \\
\bottomrule
\end{tabular}
\caption{
\textbf{Training data statistics.}
We retain 6,722 image-containing QA instances and generate one read probe per instance, balanced equally across image and text modalities.
The same QA instances are used for Read SFT, Use GRPO, and Use SFT.
}
\label{tab:training_data_stats}
\end{table}

\subsection{Training Details}
\label{app:training_details}

\subsubsection{Training Overview}
We train Qwen3-VL-4B-Instruct and Qwen3-VL-8B-Instruct in two sequential stages, initializing Read SFT from the original instruction-tuned models and Use GRPO from their respective Read SFT checkpoints.
Both stages employ Low-Rank Adaptation (LoRA) on the attention and MLP modules of the language backbone, while keeping the vision encoder and multimodal merger frozen.
This design focuses adaptation on how the language backbone processes multimodal evidence while preserving pretrained visual representations and reducing training cost.
We use a LoRA rank of 16, an alpha of 32, and dropout of 0.05 for both models.
All training uses bfloat16 precision with gradient checkpointing.
The complete training configurations for both models and stages are summarized in Table~\ref{tab:training_hyperparams}.

\subsubsection{Read SFT Training}
Read SFT optimizes the models using the selected read-probe responses described in Appendix~\ref{app:read_sft}.
We train both models for two epochs using the AdamW optimizer with a learning rate of $1\times10^{-4}$, a cosine learning rate schedule, and a warmup ratio of 0.03.
The effective batch size is 32, and the maximum sequence length is 16,384 tokens.
Checkpoints are saved at the end of each epoch.

\subsubsection{Use GRPO Training}
Use GRPO initializes from the corresponding merged Read SFT checkpoint and optimizes the models using original downstream questions with pixel-based Gold Evidence.
For both models, each rollout batch consists of 60 questions with eight sampled responses per question, yielding 480 completions.
Responses are generated with a temperature of 1.0, top-$p$ of 1.0, and a maximum completion length of 512 tokens.
We employ the DAPO loss with token-level normalization and group-standard-deviation-based advantage scaling.
Optimization uses a learning rate of $2\times10^{-5}$ with a constant schedule and five warmup steps, PPO-style policy clipping with $\epsilon=0.2$, and no KL penalty.
We additionally apply token-level importance-sampling correction for vLLM-generated rollouts, with a maximum correction ratio of 2.0.
Reward computation uses a separate gpt-oss-20b judge with temperature 0 and low reasoning effort, as detailed in Appendix~\ref{app:use_grpo}.
Both models are trained for 112 optimization steps, with checkpoints saved every 14 steps.

\subsubsection{Checkpoint Selection}
For Read SFT, we use the final checkpoint after two training epochs to initialize Use GRPO.
For Use GRPO, we independently select the checkpoint with the highest downstream QA accuracy on the validation set using Gold Evidence for each model and training variant.
All test results are reported using the selected checkpoints.

\begin{table}[t]
\centering
\renewcommand{\arraystretch}{1.12}
\setlength{\tabcolsep}{5pt}
\resizebox{\columnwidth}{!}{
\begin{tabular}{@{}lcc@{}}
\toprule
\textbf{Hyperparameter}
& \textbf{Read SFT}
& \textbf{Use GRPO} \\
\midrule
\multicolumn{3}{@{}l}{\emph{Shared Configuration}} \\
LoRA rank / alpha
& 16 / 32 & 16 / 32 \\
LoRA dropout
& 0.05 & 0.05 \\
LoRA target
& Attention + MLP & Attention + MLP \\
Vision encoder / merger
& Frozen & Frozen \\
Precision
& BF16 & BF16 \\
Gradient checkpointing
& Yes & Yes \\
Optimizer
& AdamW (fused) & AdamW (fused) \\
Weight decay
& 0 & 0 \\
Max. gradient norm
& 1.0 & 1.0 \\
Seed
& 42 & 42 \\
\midrule
\multicolumn{3}{@{}l}{\emph{Optimization}} \\
Initialization
& Original VLM & Read SFT \\
Learning rate
& $1\times10^{-4}$ & $2\times10^{-5}$ \\
LR schedule
& Cosine & Constant \\
Warmup
& 3\% & 5 steps \\
Training duration
& 2 epochs & 112 steps \\
Optimization steps (4B / 8B)
& 420 & 112 \\
Checkpoint interval
& Every epoch & Every 14 steps \\
Maximum sequence length
& 16,384 & -- \\
Effective batch size
& 32 & -- \\
Loss type
& SFT & DAPO \\
\midrule
\multicolumn{3}{@{}l}{\emph{GRPO Configuration}} \\
Questions per rollout batch
& -- & 60 \\
Responses per question
& -- & 8 \\
Completions per rollout batch
& -- & 480 \\
Steps per generation (4B / 8B)
& -- & 8 / 16 \\
Policy clipping $\epsilon$
& -- & 0.2 \\
KL coefficient
& -- & 0 \\
Advantage scaling
& -- & Group std \\
Policy iterations
& -- & 1 \\
vLLM IS correction
& -- & Token truncate (max 2.0) \\
Sampling temperature
& -- & 1.0 \\
Top-$p$
& -- & 1.0 \\
Max. completion tokens
& -- & 512 \\
\bottomrule
\end{tabular}
}
\caption{
\textbf{Training hyperparameters for Qwen3-VL-4B and Qwen3-VL-8B.}
Both models share the same configurations unless otherwise specified.
}
\label{tab:training_hyperparams}
\vspace{-0.45cm}
\end{table}

\subsection{Computational Cost}
Training was conducted primarily on four NVIDIA A100 80GB GPUs.
A \ours training run required approximately 7 (9) GPU-hours for Read SFT and 86 (106) GPU-hours for Use GRPO at the 4B (8B) scale.
Use GRPO employed four GPUs, including one hosting the gpt-oss-20b reward judge.
Across all experiments, including baselines, inference, development, and reruns, the total allocated GPU time was approximately 1,176 GPU-hours.
Offline evaluation and analysis additionally involved approximately 1.58 million GPT-6-Luna judge calls.

\section{Confidence Interval Analysis}
\label{app:95_ci}
\paragraph{Individual Confidence Intervals.}
Table~\ref{tab:main_results_ci} reports the 95\% confidence intervals (CIs) for the main results in Table~\ref{tab:main_results}.
We compute CIs using question-level percentile bootstrap resampling~\cite{efron1994introduction}, with 10,000 bootstrap replicates and a fixed random seed of 0.
For each evaluation setting, we resample questions with replacement and obtain the 95\% CI from the 2.5th and 97.5th percentiles of the resulting accuracy distribution.
For Cross-system Avg., we jointly resample the same question IDs across the three retrieved memory systems and recompute their arithmetic mean in each replicate, preserving question-level correspondence across systems.
Across both model scales, \ours achieves the highest cross-system average point estimates among the compared methods.
These results suggest that the benefits of \ours extend across heterogeneous memory systems, supporting its cross-system generalization capability.
\paragraph{Paired Confidence Intervals.}
While Table~\ref{tab:main_results_ci} reports confidence intervals for individual model accuracies, Table~\ref{tab:paired_bootstrap} assesses the differences between \ours and competing methods using paired bootstrap resampling~\cite{koehn-2004-statistical}. 
\ours significantly outperforms Use GRPO at both model scales and Use SFT $\rightarrow$ Use GRPO.

\begin{table*}[t]
\centering
\scriptsize
\setlength{\tabcolsep}{9pt}
\renewcommand{\arraystretch}{1.15}
\begin{tabular}{lccccc}
\toprule
\multirow{3}{*}{\textbf{Model / Method}}
& \multirow{3}{*}{\textbf{Gold Evidence}}
& \multicolumn{4}{c}{\textbf{Retrieved Memory Systems}} \\
\cmidrule(lr){3-6}
& & \textbf{AUGUSTUS}~\cite{jain2025augustus}
& \textbf{Omni-SimpleMem}~\cite{liu2026omni}
& \textbf{UniversalRAG}~\cite{yeo2026universalrag}
& \textbf{Cross-system Avg.} \\
\midrule

\multicolumn{6}{l}{\textit{General-purpose VLMs}} \\
GPT-5-mini
& [71.3, 75.1] & [58.9, 63.1] & [60.4, 64.5] & [61.7, 65.8] & [60.7, 64.2] \\
Qwen3-VL-32B-Instruct
& [69.4, 73.3] & [55.9, 60.1] & [56.8, 61.0] & [59.1, 63.3] & [57.6, 61.2] \\

\midrule
\multicolumn{6}{l}{\textit{Qwen3-VL-4B-Instruct}} \\
Base
& [60.9, 65.0] & [50.0, 54.3] & [50.3, 54.6] & [51.4, 55.7] & [50.9, 54.5] \\
Use SFT
& [62.0, 66.1] & [52.7, 57.0] & [53.1, 57.5] & [54.0, 58.2] & [53.6, 57.3] \\
Use GRPO
& [67.6, 71.5] & [53.4, 57.7] & [55.0, 59.2] & [53.2, 57.5] & [54.2, 57.8] \\
Use SFT $\rightarrow$ Use GRPO
& [70.6, 74.4] & [54.6, 58.8] & [56.1, 60.3] & [54.9, 59.0] & [55.5, 59.1] \\
OPD-V (4B)
& [60.8, 65.0] & [50.4, 54.7] & [52.7, 57.0] & [52.0, 56.2] & [52.1, 55.6] \\
MultiHopSpatial (4B)
& [62.0, 66.1] & [50.6, 54.8] & [53.0, 57.3] & [51.3, 55.6] & [52.0, 55.6] \\
\rowcolor{gray!10}
\textbf{\ours}
& [71.3, 75.2] & [54.3, 58.5] & [57.9, 62.1]
& [56.7, 60.9] & [56.6, 60.2] \\

\midrule
\multicolumn{6}{l}{\textit{Qwen3-VL-8B-Instruct}} \\
Base
& [60.3, 64.5] & [50.3, 54.6] & [52.4, 56.7] & [52.4, 56.7] & [52.0, 55.7] \\
Use SFT
& [64.7, 68.8] & [52.0, 56.2] & [53.7, 58.0] & [53.1, 57.5] & [53.2, 57.0] \\
Use GRPO
& [71.8, 75.6] & [55.0, 59.3] & [55.1, 59.4] & [55.8, 60.1] & [55.7, 59.3] \\
Use SFT $\rightarrow$ Use GRPO
& [71.4, 75.3] & [56.0, 60.2] & [56.1, 60.4] & [57.8, 62.0] & [57.0, 60.6] \\
OPD-V (8B)
& [62.5, 66.6] & [51.8, 56.1] & [52.7, 56.9] & [54.0, 58.2] & [53.1, 56.8] \\
MultiHopSpatial (8B)
& [64.8, 68.9] & [53.3, 57.5] & [55.2, 59.5] & [55.8, 60.1] & [55.1, 58.8] \\
\rowcolor{gray!10}
\textbf{\ours}
& [72.3, 76.1] & [58.7, 62.9] & [57.7, 62.0]
& [58.8, 63.0] & [58.7, 62.3] \\
\bottomrule
\end{tabular}
\caption{
\textbf{95\% confidence intervals for the main results on WorldMemArena (WMA).}
We report the 95\% CIs of QA accuracy (\%) under Gold Evidence and three retrieved memory systems, computed using 10,000 question-level percentile bootstrap replicates.
Cross-system Avg. denotes the arithmetic mean across the three retrieved memory systems, with question IDs jointly resampled across systems.
}
\label{tab:main_results_ci}
\vspace{-0.45cm}
\end{table*}

\begin{table}[t]
\centering
\small
\setlength{\tabcolsep}{4pt}
\renewcommand{\arraystretch}{1.15}
\begin{tabular}{lcc}
\toprule
\textbf{$\Delta$ Comparison} & \textbf{4B} & \textbf{8B} \\
\midrule
Base
& $+5.7$ {\tiny [4.3, 7.1]}
& $+6.7$ {\tiny [5.3, 8.0]} \\
Use GRPO
& $+2.4$ {\tiny [1.3, 3.6]}
& $+3.1$ {\tiny [1.9, 4.2]} \\
Use SFT $\rightarrow$ Use GRPO
& $+1.1$ {\tiny [0.1, 2.3]}
& $+1.8$ {\tiny [0.6, 2.9]} \\
\midrule
Qwen3-VL-32B
& $-1.0$ {\tiny [-2.3, 0.3]}
& $+1.1$ {\tiny [0.2, 2.4]} \\
GPT-5-mini
& $-4.0$ {\tiny [-5.5, -2.6]}
& $-1.9$ {\tiny [-3.4, -0.5]} \\
\bottomrule
\end{tabular}
\caption{
\textbf{Paired bootstrap comparison on the cross-system average.}
Each entry reports the accuracy difference
(\ours minus baseline, in percentage points) with its 95\% confidence interval from 10,000 paired bootstrap replicates.
Intervals excluding zero indicate statistically detectable differences.
}
\label{tab:paired_bootstrap}
\end{table}

\section{More Ablation Studies}
\label{app:more_ablations}
\subsection{Full result of memory}
\label{sec:detailed_memory_systems}
Table~\ref{tab:detailed_memory_systems} breaks down the main results across six reasoning categories and three retrieved memory systems.
For the 4B model, \ours achieves the highest overall accuracy on Omni-SimpleMem and UniversalRAG, while remaining competitive with the strongest baseline on AUGUSTUS.
For the 8B model, \ours consistently outperforms all compared methods in overall accuracy across all three memory systems.
At the category level, the improvements are particularly evident in certain reasoning categories, including commonsense and knowledge-based reasoning, although the gains vary across memory systems.
These results further demonstrate the effectiveness of \ours across heterogeneous memory systems while highlighting differences in performance across reasoning categories.

\subsection{Counterfactual Memory Utilization}
\label{sec:counterfactual_memory}
\begin{figure*}[t]
    \centering
    \begin{tabular}{
        @{}l
        @{\hspace{0.025\textwidth}}
        p{0.32\textwidth}
        @{\hspace{0.04\textwidth}}
        p{0.32\textwidth}@{}
    }
        \toprule
        & \centering\textbf{Original Memory}
        & \centering\arraybackslash\textbf{Counterfactual Memory} \\
        \midrule

        \textbf{Text}
        &
        \parbox[c][4.5em][c]{\linewidth}{
            \small\raggedright
            ``...the agent clicked the
            \textcolor{blue}{\textbf{VSCode icon}}
            in the taskbar.''
        }
        &
        \parbox[c][4.5em][c]{\linewidth}{
            \small\raggedright
            ``...the agent clicked the
            \textcolor{red}{\textbf{Chrome icon}}
            in the taskbar.''
        }
        \\[10pt]
        \midrule

        \textbf{Pixel}
        &
        \centering
        \includegraphics[width=0.8\linewidth]{
            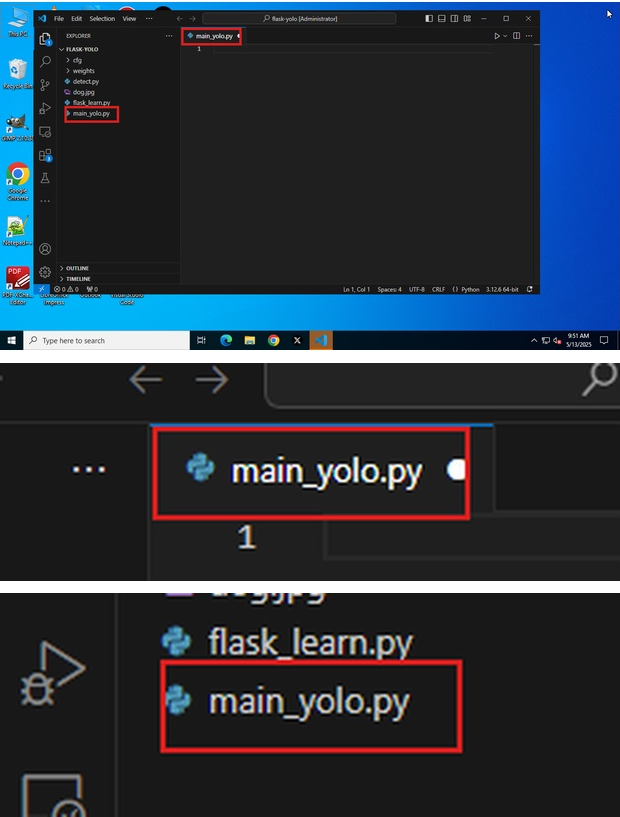
        }
        &
        \centering\arraybackslash
        \includegraphics[width=0.8\linewidth]{
            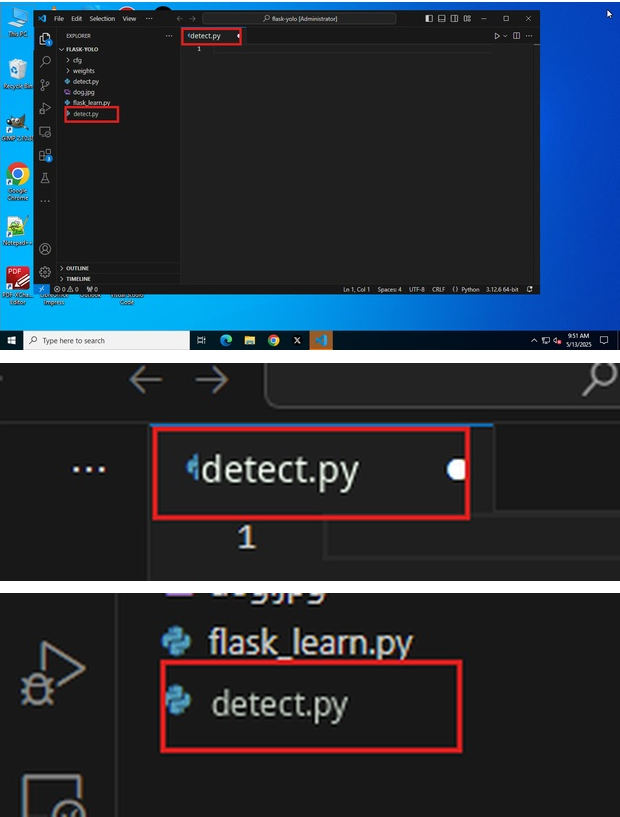
        }
        \\
        \bottomrule
    \end{tabular}

    \caption{
        \textbf{Examples of counterfactual memory interventions.}
        In textual memory, the application icon is changed
        from \textit{VSCode} to \textit{Chrome}.
        In visual memory, the filename
        \texttt{main\_yolo.py} is replaced with
        \texttt{detect.py} directly in the image pixels.
        The original questions remain unchanged.
    }
    \label{fig:counterfactual_example}
\end{figure*}

\begin{table}[t]
    \centering
    \resizebox{\columnwidth}{!}{
    \begin{tabular}{llcccc}
        \toprule
        Metric & Edit ($N$) & Base
        & Use SFT $\rightarrow$ GRPO
        & MemRU & $\Delta$ ($p$) \\
        \midrule
        \multirow{2}{*}{Probe Follow}
        & Pixel (115) & \textbf{88.7} & 86.1 & 87.8
        & +1.7 (.63) \\
        & Text (732) & 88.5 & 90.2 & \textbf{91.1}
        & +1.0 (.27) \\
        \midrule
        \multirow{2}{*}{Use Follow}
        & Pixel (115) & 48.7 & 46.1 & \textbf{57.4}
        & \textbf{+11.3 (.007)} \\
        & Text (732) & 40.0 & 58.2 & \textbf{59.6}
        & +1.4 (.32) \\
        \bottomrule
    \end{tabular}
    }
    \caption{
        \textbf{Counterfactual memory utilization on WMA.}
        \emph{Probe Follow} measures direct evidence reading, while \emph{Use Follow} measures evidence utilization in downstream QA.
        $\Delta$ denotes MemRU minus Use SFT $\rightarrow$ GRPO
        (paired McNemar $p$-values).
    }
    \label{tab:counterfactual_memory}
\end{table}
To further examine whether improved downstream QA performance reflects effective memory utilization, we conduct counterfactual interventions on WorldMemArena by replacing an answer-critical fact in the oracle memory with a plausible alternative while keeping the original question unchanged.
Two intervention types are considered: modifying textual memory and directly editing text rendered in memory images, as illustrated in Figure~\ref{fig:counterfactual_example}.
For textual interventions, the original value is replaced with an alternative in the memory text.
For pixel-level interventions, OCR is first applied to locate the answer-critical text within memory images.
The detected text region is then erased using the estimated background color, and the alternative value is rendered at the same location with an adjusted font size to match the original text width.
The evaluation compares the base model, Use SFT $\rightarrow$ Use GRPO, and MemRU at the 4B scale, measuring \emph{Probe Follow}, which assesses whether the modified fact can be directly recovered, and \emph{Use Follow}, which measures whether the original downstream answer reflects the modified evidence.
As shown in Table~\ref{tab:counterfactual_memory}, MemRU achieves a substantially higher Use Follow rate on pixel-level interventions than Use SFT $\rightarrow$ Use GRPO (57.4\% vs.\ 46.1\%), despite comparable Probe Follow rates (87.8\% vs.\ 86.1\%).
In contrast, both methods achieve similar Use Follow rates on textual interventions (59.6\% vs.\ 58.2\%).
These results suggest that MemRU more effectively incorporates evidence from memory images into downstream answers, even when the relevant information can be recovered at similar rates, providing intervention-based support for improved memory utilization.

\begin{table*}[t]
\centering
\setlength{\tabcolsep}{3pt}
\renewcommand{\arraystretch}{1.12}
\resizebox{\textwidth}{!}{
\begin{tabular}{cl*{21}{c}}
\toprule
\multirow{2}{*}{\textbf{Model}}
& \multirow{2}{*}{\textbf{Method}}
& \multicolumn{7}{c}{\textbf{AUGUSTUS}}
& \multicolumn{7}{c}{\textbf{Omni-SimpleMem}}
& \multicolumn{7}{c}{\textbf{UniversalRAG}} \\
\cmidrule(lr){3-9}
\cmidrule(lr){10-16}
\cmidrule(lr){17-23}
&
& VFR & VU & CMR & TR & KR & FR & Overall
& VFR & VU & CMR & TR & KR & FR & Overall
& VFR & VU & CMR & TR & KR & FR & Overall \\
\midrule
\multirow{7}{*}{\textbf{4B}}
& Base
& 60.2 & 54.8 & 44.9 & 20.8 & 52.2 & 78.3 & 52.1
& 61.4 & 53.7 & 43.7 & 20.2 & 54.2 & 80.0 & 52.4
& 63.9 & 56.3 & 47.4 & 19.4 & 53.9 & 79.2 & 53.5 \\
& Use SFT
& 59.4 & 57.5 & 48.9 & 22.5 & 61.2 & 78.0 & 54.8
& 55.4 & 61.9 & 46.4 & 24.6 & 62.1 & 80.0 & 55.3
& 61.1 & 63.0 & 49.5 & 22.0 & 60.1 & 79.4 & 56.1 \\
& Use GRPO
& 61.1 & 57.2 & 48.3 & 24.6 & 62.6 & 78.3 & 55.6
& 59.9 & 60.7 & 48.3 & 23.7 & 64.3 & 84.2 & 57.1
& 67.0 & 56.6 & 48.6 & 20.5 & 59.3 & 78.6 & 55.3 \\
& Use SFT $\rightarrow$ Use GRPO
& 62.2 & 61.3 & 49.8 & 25.1 & 61.8 & 78.9 & \textbf{56.7}
& 60.8 & 61.9 & 52.3 & 24.3 & 68.3 & 80.0 & 58.1
& 62.5 & 62.8 & 48.9 & 21.7 & 66.6 & 78.0 & 57.0 \\
& OPD-V
& 62.2 & 50.4 & 46.4 & 22.5 & 55.6 & 76.6 & 52.5
& 58.2 & 56.0 & 48.3 & 23.1 & 58.7 & 83.4 & 54.8
& 64.8 & 56.9 & 46.4 & 21.1 & 54.5 & 79.7 & 54.1 \\
& MultiHopSpatial
& 60.2 & 53.1 & 46.1 & 23.4 & 56.2 & 76.1 & 52.7
& 59.4 & 54.8 & 48.6 & 22.8 & 61.0 & 83.1 & 55.2
& 63.9 & 53.7 & 46.4 & 21.7 & 55.3 & 78.3 & 53.4 \\
\rowcolor{gray!10}
& \textbf{\ours}
& 63.6 & 59.2 & 50.8 & 23.1 & 61.8 & 78.6 & 56.4
& 64.8 & 63.6 & 54.2 & 24.0 & 68.8 & 83.4 & \textbf{60.0}
& 67.9 & 65.4 & 50.2 & 23.4 & 63.5 & 81.1 & \textbf{58.8} \\

\midrule

\multirow{8}{*}{\textbf{8B}}
& Base
& 56.5 & 56.3 & 41.8 & 22.8 & 45.2 & 78.0 & 52.4
& 53.4 & 53.7 & 47.4 & 27.7 & 49.2 & 81.1 & 54.6
& 54.8 & 56.3 & 45.2 & 28.6 & 49.7 & 78.9 & 54.5 \\
& Use SFT
& 59.1 & 61.6 & 47.1 & 24.0 & 54.5 & 77.5 & 54.1
& 53.7 & 65.4 & 45.8 & 27.5 & 61.2 & 80.0 & 55.8
& 57.1 & 62.8 & 44.6 & 27.5 & 60.1 & 78.0 & 55.2 \\
& Use GRPO
& 64.5 & 55.4 & 51.1 & 26.6 & 62.6 & 81.1 & 57.1
& 59.9 & 59.5 & 46.7 & 26.3 & 64.6 & 84.8 & 57.3
& 65.3 & 59.2 & 47.4 & 30.3 & 64.3 & 79.4 & 57.9 \\
& Use SFT $\rightarrow$ Use GRPO
& 61.6 & 64.2 & 51.5 & 26.9 & 61.8 & 78.3 & 58.0
& 56.8 & 63.6 & 49.0 & 26.3 & 67.1 & 80.6 & 58.2
& 63.6 & 65.1 & 51.8 & 26.6 & 70.2 & 78.0 & 59.9 \\
& OPD-V
& 61.1 & 56.6 & 47.1 & 21.1 & 55.3 & 81.1 & 53.9
& 59.4 & 58.5 & 45.5 & 22.5 & 56.7 & 84.5 & 54.8
& 61.4 & 57.1 & 50.5 & 23.7 & 59.6 & 83.1 & 56.1 \\
& MultiHopSpatial
& 61.6 & 58.1 & 47.1 & 26.3 & 58.7 & 79.2 & 55.4
& 59.9 & 56.5 & 53.3 & 25.7 & 64.9 & 82.8 & 57.4
& 63.4 & 58.8 & 53.9 & 26.6 & 63.8 & 80.0 & 57.9 \\
\rowcolor{gray!10}
& \textbf{\ours}
& 64.8 & 64.2 & 54.8 & 30.1 & 67.1 & 82.5 & \textbf{60.8}
& 61.1 & 64.8 & 51.4 & 27.2 & 69.7 & 83.7 & \textbf{59.9}
& 64.2 & 65.1 & 48.9 & 30.3 & 72.5 & 82.3 & \textbf{60.8} \\

\midrule

\multirow{2}{*}{\textbf{Reference}}
& Qwen3-VL-32B
& 64.8 & 58.9 & 49.8 & 29.2 & 62.9 & 80.8 & 58.0
& 62.8 & 60.7 & 54.8 & 24.3 & 65.4 & 84.2 & 58.9
& 65.9 & 63.3 & 52.6 & 32.1 & 67.7 & 84.2 & 61.2 \\
& GPT-5-mini
& 69.6 & 56.9 & 53.6 & 32.4 & 70.5 & 82.0 & 61.1
& 67.9 & 57.8 & 53.9 & 31.5 & 75.3 & 86.8 & 62.5
& 71.0 & 60.1 & 55.7 & 37.3 & 71.9 & 84.8 & 63.7 \\

\bottomrule
\end{tabular}
}
\caption{
\textbf{Detailed performance across memory systems and reasoning categories.}
QA accuracy (\%) across six reasoning categories on AUGUSTUS, Omni-SimpleMem, and UniversalRAG.
}
\label{tab:detailed_memory_systems}
\vspace{-0.3cm}
\end{table*}

\end{document}